\documentclass{article}
\usepackage[preprint]{tmlr}
\usepackage{amsmath,amssymb,amsthm}
\usepackage{mathtools}
\usepackage{algorithm}
\usepackage{algpseudocode}
\usepackage{booktabs}
\usepackage{multirow}
\usepackage{hyperref}
\usepackage{cleveref}
\usepackage{bm}
\usepackage{xcolor}
\usepackage{enumitem}
\usepackage{microtype}
\usepackage{graphicx}
\usepackage{tikz}
\usetikzlibrary{positioning, arrows.meta, shapes.geometric, backgrounds, fit, calc, decorations.pathreplacing, decorations.pathmorphing}

\newtheorem{proposition}{Proposition}
\newtheorem{theorem}{Theorem}

\DeclareMathOperator{\E}{\mathbb{E}}
\DeclareMathOperator{\tr}{tr}

\newcommand{\R}{\mathbb{R}}
\newcommand{\cS}{\mathcal{S}}

\newcommand{\cL}{\mathcal{L}}
\newcommand{\abar}{\bar{\mathbf{a}}}
\newcommand{\sighat}{\hat{\sigma}}
\newcommand{\SigV}{\bm{\Sigma}_V}

\title{L\'evy Attention: Single-Pass Predictive Uncertainty\\ for Continuous-Time Attention}

\author{%
  \name Sotirios P. Chatzis \email sotirios.chatzis@cut.ac.cy \\
  \addr GenML Laboratory, Department of Electrical Engineering \\
  \addr and Computer Science and Engineering \\
  \addr Cyprus University of Technology, Limassol, Cyprus
  \AND
  \name Loukas Papadoulas \email luke@ethicalaicy.com \\
  \addr Ethical AI Novelties \\
  \addr Cyprus
}

\def\month{08}
\def\year{2026}
\def\openreview{https://openreview.net/forum?id=XXXX}

\begin{document}
\maketitle

\begin{abstract}
Deep models for irregularly-sampled time series answer queries at arbitrary continuous timestamps, yet report nothing about how far each answer should be trusted. We show that the attention layer itself can close that gap: with the right stochastic formulation, the pass that makes each prediction also reports, in closed form and at no extra cost, how far that prediction should be trusted. We introduce \textbf{L\'evy Attention}, a cross-attention operator whose output is a stochastic integral against an inhomogeneous Poisson random measure: query--key compatibilities assemble an intensity over a continuous (time $\times$ channel) index space, the measure scatters atoms under it, and the output averages an interpolated value field at those atoms. In expectation L\'evy Attention reduces to a mollified cosine-kernel attention, so it replaces a softmax layer and trains with exact gradients.
What distinguishes it is what softmax discards: row normalisation divides out the total compatibility mass, and the spread of the attended values is never surfaced. The Poisson construction preserves both in closed form as the \emph{evidence} $\Lambda_q$ and the \emph{disagreement} $\tr\SigV(q)$, and an exact variance identity makes their combination $\sighat(q)=\sqrt{\smash[b]{\tr\SigV(q)\,\varphi(\Lambda_q)}}$ the root-mean-square deviation of the sampled operator. No head is trained for it: the deterministic pass that makes the prediction emits it.
In the experiments, disagreement carries the signal; the evidence factor swings from uninformative on dense data to strongly informative on sparse. On the t-PatchGNN benchmark the operator swap costs at most $5.6\%$ accuracy against a matched control and nothing on the sparsest dataset, and a transplant into the benchmark's own model preserves its accuracy. Across matched five-seed suites the free disagreement signal improves on the sparsification error of 20-pass MC dropout in every seed of a dense and a sparse benchmark, and attains the best such error of any training-free estimator we measure on the dense one. The combined $\sighat$ scales a validation-calibrated Gaussian whose zero-sample CRPS improves on a fifty-draw sampler, a split-conformal wrapper on top of it reaches nominal coverage at every level, and one pass ranks an unseen cohort of 3{,}383 patients by trust in 1.4 seconds.
\end{abstract}

\section{Introduction}
\label{sec:intro}

Models for irregularly-sampled multivariate time series (IMTS), such as clinical records, climate stations, and asynchronous sensors, increasingly follow a common pattern: encode the observed $(t_i, \text{variable}, \text{value})$ triplets, then answer queries at \emph{arbitrary continuous timestamps} through attention \citep{shukla2021mtan,zhang2024tpatchgnn,yalavarthi2024grafiti}. The prediction interface is a function of continuous time; the natural next question is a per-query one: \emph{how much should this particular answer be trusted?} A vital-sign estimate interpolated inside a dense observation window and one extrapolated across a six-hour gap in a sparse record should not carry the same authority. Downstream consumers (alarm thresholds, active sensing, compute triage) need to know the difference.

The standard answers all pay for uncertainty separately from prediction. Deep ensembles \citep{lakshminarayanan2017} train and evaluate $K$ models; MC dropout \citep{gal2016} runs $K$ stochastic passes; diffusion-based imputers such as CSDI \citep{tashiro2021csdi} draw ${\sim}100$ samples per query; learned variance or confidence heads \citep{kendall2017} add a trained output and pay for it in accuracy (\S\ref{sec:exp_unc}). A parallel line on \emph{single-pass} deterministic uncertainty \citep{liu2020sngp,mukhoti2023ddu} obtains uncertainty from one forward pass, but does so by constraining or augmenting the network around a standard backbone. None of these derives uncertainty from the attention mechanism itself, the very component that decides, query by query, which observations the answer rests on.

This paper takes that step: it turns the attention layer itself into the uncertainty interface, at zero marginal cost --- no second model, no extra passes, no trained head, no change to the loss. We formulate cross-attention over a continuous index space as \emph{stochastic integration against an inhomogeneous Poisson random measure}:\footnote{The name \emph{L\'evy} records that a Poisson random measure is the jump mechanism of a L\'evy process (the L\'evy--It\^o decomposition); the output $\int V\,dN_q$ is the corresponding compound-Poisson integral \citep{kingman1993}.} the query--key compatibility kernel defines an intensity $\lambda_q(s)$ over the index space, a Poisson measure $N_q$ scatters random atoms under that intensity, and the output averages a continuously interpolated value field at the atoms.

\begin{figure*}[!tp]
\centering
\resizebox{0.80\textwidth}{!}{
\begin{tikzpicture}[
    >=Stealth,
    node distance=1.5cm and 2cm,
    font=\small,
    box/.style={rectangle, draw=black!70, thick, rounded corners, fill=white, align=center, minimum height=1cm},
    token/.style={rectangle, draw=black!50, fill=blue!10, minimum size=0.6cm, font=\ttfamily},
    query/.style={rectangle, draw=red!60, thick, fill=red!10, minimum size=0.6cm, font=\ttfamily},
    axis/.style={->, thick, black!70},
    curve/.style={thick, blue!80},
    sstar/.style={star, star points=5, star point ratio=2.2, fill=orange!90!black, draw=orange!50!black, line width=0.25pt, minimum size=7pt, inner sep=0pt},
]
\useasboundingbox (-0.3, -6.8) rectangle (16.6, 3.8);

\node[query] (Q) {$q$};
\node[token, below=1cm of Q] (K1) {$k_1$};
\node[token, right=0.3cm of K1] (K2) {$k_2$};
\node[above=0.1cm of K2] {\dots};
\node[token, right=0.8cm of K2] (Kn) {$k_n$};

\node[token, below=0.6cm of K1, fill=green!10] (V1) {$v_1$};
\node[token, below=0.6cm of K2, fill=green!10] (V2) {$v_2$};
\node[above=0.1cm of V2] {\dots};
\node[token, below=0.6cm of Kn, fill=green!10] (Vn) {$v_n$};

\node[font=\bfseries, anchor=west] at (-0.3, 1.1) {Stage 1};
\node[font=\scriptsize, anchor=west, align=left] at (-0.3, 0.72) {positions from\\ timestamps};

\coordinate (S_start) at (3.5, 0);
\coordinate (S_end) at (10.5, 0);

\draw[axis] (S_start) -- (S_end) node[below right, font=\scriptsize] {time $\in[0,1]$};
\node[font=\bfseries, anchor=west] at (3.5, 3.5) {Stages 2--3: intensity \& Poisson measure $N_q$};

\draw[->, dashed, black!60] (K1.east) to[out=0, in=180] node[midway, below, font=\scriptsize] {$t_i$ (given)} (4.2, 0);
\draw[->, dashed, black!60] (K2.east) to[out=0, in=180] (5.6, 0);
\draw[->, dashed, black!60] (Kn.east) to[out=0, in=180] (9.6, 0);

\draw[curve, fill=blue!5]
    (3.5,0) -- (3.8,0.1)
    .. controls (4.0,0.5) and (4.1,1.8) .. (4.2,1.8)
    .. controls (4.3,1.8) and (4.4,0.2) .. (4.6,0.1)
    -- (5.2,0.1)
    .. controls (5.4,0.3) and (5.5,1.0) .. (5.6,1.0)
    .. controls (5.7,1.0) and (5.8,0.2) .. (6.0,0.1)
    -- (8.8,0.05)
    .. controls (9.3,0.2) and (9.5,1.5) .. (9.6,1.5)
    .. controls (9.7,1.5) and (9.8,0.1) .. (10.1,0.05)
    -- (10.4,0);

\node[blue!80] at (7.3, 2.75) {$\lambda_q(s) = \tau \sum_i \kappa(q, k_i)\,\delta_\varepsilon(s - s_i)$};

\draw[->, thick, black!65, decorate, decoration={snake, amplitude=0.3mm, segment length=3mm, post length=2mm}] (4.2, 1.7) -- (4.2, 0.3);
\draw[->, thick, black!65, decorate, decoration={snake, amplitude=0.3mm, segment length=3mm, post length=2mm}] (5.6, 0.9) -- (5.6, 0.3);
\draw[->, thick, black!65, decorate, decoration={snake, amplitude=0.3mm, segment length=3mm, post length=2mm}] (9.6, 1.4) -- (9.6, 0.3);

\filldraw[red!80] (4.12, 0) circle (1.8pt);
\node[below, yshift=-2pt, font=\tiny, red!70!black] at (4.12, 0) {$S_1$};
\filldraw[red!80] (4.25, 0) circle (1.8pt);
\node[below, yshift=-8pt, font=\tiny, red!70!black] at (4.25, 0) {$S_2$};
\filldraw[red!80] (5.6, 0) circle (1.8pt);
\node[below, yshift=-2pt, font=\tiny, red!70!black] at (5.6, 0) {$S_3$};
\filldraw[red!80] (9.6, 0) circle (1.8pt);
\node[below, yshift=-2pt, font=\tiny, red!70!black] at (9.6, 0) {$S_{Z_q}$};

\node[sstar] at (4.2, 1.95) {};
\node[sstar] at (5.6, 1.15) {};
\node[sstar] at (9.6, 1.65) {};
\node[font=\tiny, orange!70!black] at (4.2, 2.18) {$s_1$};
\node[font=\tiny, orange!70!black] at (5.6, 1.38) {$s_2$};
\node[font=\tiny, orange!70!black] at (9.6, 1.88) {$s_n$};

\draw[<->, thick, black!50] (6.4, -0.45) -- (8.5, -0.45);
\node[black!65, font=\tiny, align=center] at (7.45, -0.75) {$\lambda_q \approx 0 \Rightarrow$ no atoms};

\coordinate (V_start) at (3.5, -4.5);
\coordinate (V_end) at (10.5, -4.5);
\draw[axis] (V_start) -- (V_end) node[below right, font=\scriptsize] {time $\in[0,1]$};
\node[font=\bfseries, anchor=west] at (3.5, -2.85) {Stage 4: value field $V(s)$ (RBF interpolation)};

\draw[->, dashed, black!40] (V1.east) to[out=0, in=180] (4.2, -4.5);
\draw[->, dashed, black!40] (V2.east) to[out=0, in=180] (5.6, -4.5);
\draw[->, dashed, black!40] (Vn.east) to[out=0, in=180] (9.6, -4.5);

\draw[thick, green!60!black]
    (3.5,-4.3) .. controls (4.2,-3.8) .. (5.0,-4.2)
    .. controls (5.8,-4.7) .. (7.0,-4.1)
    .. controls (8.5,-3.8) .. (9.6,-4.6)
    .. controls (10.1,-4.8) .. (10.4,-4.5);
\node[green!60!black, font=\scriptsize] at (9.0, -3.6) {$V(s)$};

\draw[dashed, red!40, ->] (4.12, -0.5) -- (4.12, -3.95);
\draw[dashed, red!40, ->] (4.25, -0.7) -- (4.25, -3.95);
\draw[dashed, red!40, ->] (5.6, -0.5) -- (5.6, -4.55);
\draw[dashed, red!40, ->] (9.6, -0.5) -- (9.6, -4.55);

\draw [decorate,decoration={brace,amplitude=6pt,mirror}, thick, red!70!black]
    (4.0,-5.25) -- (9.8,-5.25) node[midway,below=6pt, font=\small, font=\bfseries] (SumNode) {$\sum_{j=1}^{Z_q} V(S_j)$};

\node[box, draw=blue!50!black, font=\scriptsize, align=center, anchor=north] (Evid) at (14.7, 3.6)
    {$\Lambda_q=\int\lambda_q\,d\cL$\\ \textbf{evidence}: total mass\\ the query found};
\draw[->, blue!60!black, thin] (Evid.west) to[out=190, in=40] (10.0, 1.1);

\node[box, fill=blue!3, font=\scriptsize, align=center, anchor=north] (Fallback) at (14.7, 1.55)
    {$\abar(q) = \dfrac{\int V\,\lambda_q\,d\cL}{\Lambda_q}$\\[2pt] mean (training path)};
\draw[->, thick, dashed, blue!70!black] (10.4, 0.25) to[out=10, in=180] (Fallback.west);

\node[box, fill=blue!5, align=center, font=\scriptsize, anchor=north] (Output) at (14.7, -0.45)
    {$\mathbf{a}(q) = \begin{dcases}\tfrac{1}{Z_q}\!\sum_j\! V(S_j), & Z_q\geq 1 \\[2pt] \abar(q), & Z_q = 0\end{dcases}$};
\node[font=\bfseries, anchor=south] at (Output.north) {Output};
\draw[->, thick, red!70!black] (SumNode.east) to[out=0, in=220] node[pos=0.75, below right, font=\tiny] {$\div Z_q$} (Output.220);
\draw[->, thick, dashed, blue!70!black] (Fallback.south) -- (Output.north);

\node[box, draw=green!40!black, font=\scriptsize, align=center, anchor=north] (Disag) at (14.7, -2.5)
    {$\tr\SigV(q)$\\ \textbf{disagreement}: spread of\\ $V$ under the beam};
\draw[->, green!40!black, thin] (Disag.west) to[out=190, in=20] (10.2, -4.1);

\node[box, fill=orange!6, draw=orange!60!black, font=\scriptsize, align=center, anchor=north] (Sig) at (14.7, -4.6)
    {$\sighat(q)=\sqrt{\tr\SigV(q)\,\varphi(\Lambda_q)}$\\[1pt] {\tiny$\varphi(\Lambda)\simeq1/\Lambda$}\\[1pt] \textbf{uncertainty}, pre-sample};
\draw[->, thin, blue!50!black] (Evid.east) to[out=0, in=0, looseness=1.6] (Sig.east);
\draw[->, thin, green!40!black] (Disag.south) -- (Sig.north);
\end{tikzpicture}
}
\caption{The L\'evy Attention pipeline on a time series. Observation tokens sit at positions $s_i=(t_i,c_i)$ (orange stars; the channel coordinate is suppressed in the sketch). The intensity $\lambda_q(s)$ is the flashlight beam of \S\ref{sec:intro}; the Poisson measure scatters atoms under its peaks (red dots) and none where the beam is dark, and the output averages the value field there. The same pass meters the beam: total mass $\Lambda_q$ (\emph{evidence}), value spread $\tr\SigV(q)$ (\emph{disagreement}), and their combination $\sighat(q)$ (\cref{thm:variance}).}
\label{fig:concept}
\end{figure*}
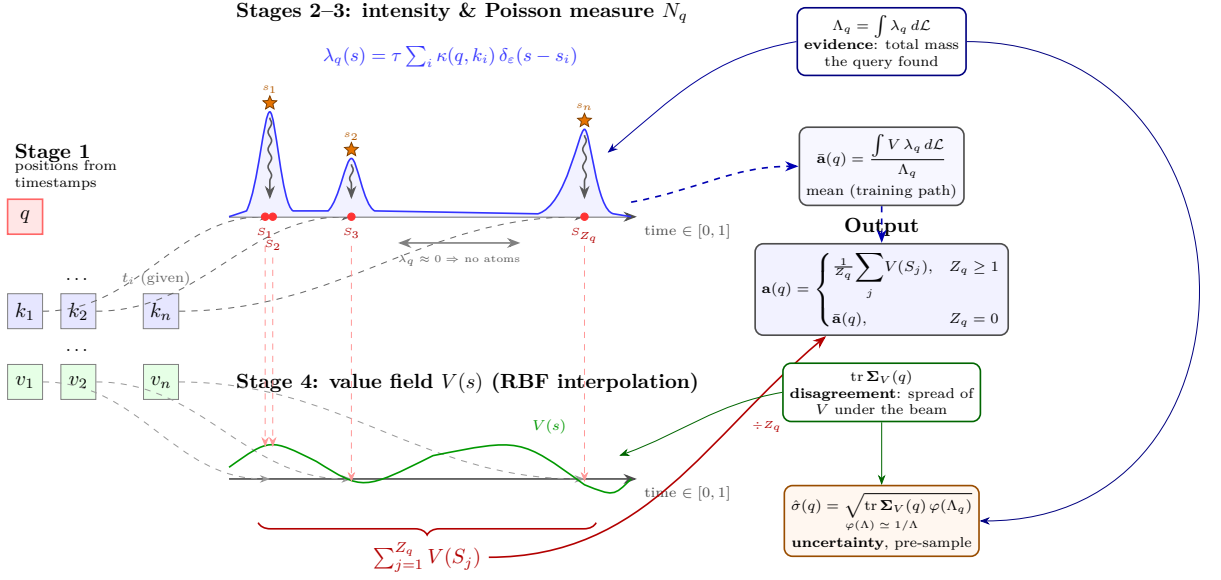

\paragraph{Intuition.} Picture the query as a flashlight sweeping the timeline of observations, shining brighter over the ones it matches well. Softmax attention averages the brightness everywhere and reports a single number. L\'evy Attention instead lets the brightness \emph{scatter a random handful of points}. More points fall where the beam is brighter; none fall where it is dark. The output averages the observed values at those points (\cref{fig:concept}). The brightness then does triple duty. Its \emph{total}, $\Lambda_q$, is how much light the query found at all: the \emph{evidence}. How far the lit values spread, counting the brightest most, is how much those observations disagree: the \emph{disagreement}, $\tr\SigV(q)$. The square root of their ratio is how much the sampled answer would wobble from draw to draw. It is readable from the beam itself, before a single point is drawn. Three properties follow from this one construction:

\begin{enumerate}[leftmargin=2em,itemsep=2pt]
\item \textbf{Drop-in mean.} In expectation L\'evy Attention reduces to a mollified cosine-kernel attention (Theorem~\ref{thm:mean}), so the deterministic mean path (\S\ref{sec:modes}) trains with exact gradients and replaces a softmax layer without architectural surgery.
\item \textbf{Closed-form, pre-sample uncertainty.} The same forward pass emits two per-query statistics in closed form: the total intensity $\Lambda_q$ (the \emph{evidence}) and the value spread $\tr\SigV(q)$ (the \emph{disagreement}). Both are short reductions over quantities the pass has already computed. Their extra cost is $O(L)$ per query on top of a shared $O(Ld)$ precompute, where $L$ is the number of cells of the discretised index space (128 by default) and $d$ is the value dimension; this is negligible next to the $O(nd)$ kernel row. \Cref{thm:variance} shows that the output deviates from its mean with root-mean-square deviation exactly $\sighat(q)=\sqrt{\tr\SigV(q)\,\varphi(\Lambda_q)}$, where $\varphi(\Lambda)\approx 1/\Lambda$. No sampling, no extra pass, no auxiliary head.
\item \textbf{Cheap error bars on demand.} When a point prediction is not enough, error bars come almost free from the same trained model. Everything expensive --- the encoder, the compatibility kernel, the beam and the values it lights --- is computed once, and one sample costs no more than scattering a fresh handful of points under that same beam and re-averaging. $K$ samples therefore cost one encoder pass plus $K$ redraws in operation count, rather than $K$ full passes, and their spread is the error bar (\S\ref{sec:modes}).
\end{enumerate}

Evidence and disagreement are the two failure modes attention can report on itself: the query may find \emph{too little} relevant mass, or it may find plenty that \emph{conflicts}. Softmax attention discards the first at the row-normalisation, dividing the partition function out, and never surfaces the second; the Poisson construction preserves both. The disagreement factor gives the better \emph{ranking} of queries; $\sighat$, which carries the units of the operator's own deviation, is what \S\ref{sec:exp_dist} calibrates into an interval scale.

\paragraph{Contributions.}
\begin{enumerate}[leftmargin=2em,itemsep=2pt]
\item \textbf{Mechanism} (\S\ref{sec:method}): attention as stochastic integration against a Poisson random measure on a (time $\times$ channel) index space, with a deterministic mean path for training and a sampling path for on-demand error bars.
\item \textbf{Theory} (\S\ref{sec:theory}): a mean identity and an \emph{exact variance identity} whose scale $\sighat(q)$ factorises into disagreement over evidence; we also bound the mollification bias.
\item \textbf{Evidence} (\S\ref{sec:experiments}), organised around one claim, \emph{the uncertainty is free}. Accuracy survives the swap: on the sparsest dataset our backbone attains a lower MSE than every published value with either decode layer, which we therefore credit to the backbone family rather than to the operator, and a drop-in transplant into t-PatchGNN leaves that model's accuracy within its own seed spread. The free disagreement signal then beats 20-pass MC dropout on AUSE in every seed of matched five-seed suites on a dense and a sparse benchmark, and on the dense one matches the best ranking any frozen-model estimator attains in our suite; better ranking exists only by changing the model or fitting on labels, and we measure what each route costs. Calibrated and split-conformal intervals follow from the identity, one pass screens an unseen $3{,}383$-record cohort, and ablations tie each ingredient of the construction to its effect.
\end{enumerate}

\section{L\'evy Attention}
\label{sec:method}

\subsection{Setting and notation}
A query set $\{q_j\}_{j=1}^m$ cross-attends into $n$ observation tokens with features $k_i, v_i \in \R^{d}$ (per head) and \emph{timestamps} $t_i \in [0,1]$ (normalised time). The index space is $\cS = [0,1]^2$: the first axis is physical time, the second a learned \emph{channel coordinate} that spreads variables observed at the same instant, at the resolution the grid allows. Integrals over $\cS$ are taken with respect to the Lebesgue measure, written $\cL$. Keys receive positions
\begin{equation}
\label{eq:positions}
s_i \;=\; \bigl(t_i,\; c_i\bigr), \qquad c_i = \bigl(1+e^{-(w_v^\top x_i + b_v)}\bigr)^{-1} \in (0,1),
\end{equation}
where $x_i\in\R^{d_{\text{model}}}$ is the token's feature before the per-head split and $(w_v,b_v)$ is one affine map per head, the only learned part; the time coordinate is \emph{given by the data}.

\subsection{The operator in four steps}
\label{sec:stages}

\textbf{(i) Compatibility and intensity.} Query--key compatibility uses a scaled cosine kernel \citep{liu2022swinv2}: with $\hat q = q/\lVert q\rVert$, $\hat k_i = k_i/\lVert k_i\rVert$,
\begin{equation}
\label{eq:kernel}
\kappa(q,k_i) \;=\; \exp\!\bigl(\sqrt{d}\,\langle \hat q,\hat k_i\rangle\bigr).
\end{equation}
$\kappa$ is \emph{not} row-normalised, so its absolute scale is meaningful: it will become the evidence. The cosine form bounds the logit to $[-\sqrt d, \sqrt d]$, keeping unnormalised masses on a comparable scale across queries, where raw dot-product logits would let them vary over orders of magnitude. Each key spreads its compatibility mass over $\cS$ through a Gaussian mollifier, centred at its position and normalised to unit Lebesgue mass per key. The two axes of $\cS$ carry no shared scale, so each has its own bandwidth: for $u=(u_t,u_v)$ and $\varepsilon=(\varepsilon_t,\varepsilon_v)$, write $\lVert u\rVert_\varepsilon^2 \coloneqq u_t^2/\varepsilon_t^2 + u_v^2/\varepsilon_v^2$. Then
\begin{equation}
\label{eq:mollifier}
\delta_\varepsilon(s - s_i) \;=\; \frac{e^{-\lVert s-s_i\rVert_\varepsilon^2/2}}{\int_\cS e^{-\lVert s'-s_i\rVert_\varepsilon^2/2}\,d\cL(s')},
\qquad \int_\cS \delta_\varepsilon(s - s_i)\,d\cL(s) = 1
\end{equation}
In words, $\delta_\varepsilon$ is a smoothed stand-in for a point mass at $s_i$: a Gaussian bump that integrates to one over $\cS$, i.e.\ a probability density on the index space. Normalising over $\cS$ rather than by the closed-form Gaussian constant matters near the boundary of $[0,1]^2$, where part of the bump would otherwise fall outside the domain and the key would lose mass. On this basis, we introduce the \emph{attention intensity}
\begin{equation}
\label{eq:intensity}
\lambda_q(s) \;=\; \tau \sum_{i=1}^n \kappa(q,k_i)\,\delta_\varepsilon(s - s_i),
\qquad
\Lambda_q \;\coloneqq\; \int_\cS \lambda_q \,d\cL \;=\; \tau\sum_{i=1}^n \kappa(q,k_i),
\end{equation}
with rate $\tau>0$, fixed at $0.5$ throughout. The total intensity $\Lambda_q$ is exactly the (scaled) partition function of the kernel: the quantity softmax divides out and discards.

\textbf{(ii) Random measure.} The attention intensity now drives a sampling mechanism. $N_q$ is the Poisson random measure on $\cS$ with intensity $\lambda_q$: a random scatter of points in the index space, called \emph{atoms}, in which every region $A\subseteq\cS$ receives a Poisson-distributed number of atoms with mean $\int_A \lambda_q\,d\cL$, independently across disjoint regions. Atoms therefore concentrate under the peaks of the intensity, where the compatible observations sit. Where the query found no compatible mass, the intensity vanishes and no atoms fall. In particular, the total atom count is itself Poisson:
\begin{equation}
\label{eq:count}
Z_q \;\coloneqq\; N_q(\cS) \;\sim\; \mathrm{Poisson}(\Lambda_q).
\end{equation}
The evidence a query finds determines how many atoms it gets to average over.

\textbf{(iii) Value field.} The atoms of $N_q$ fall at arbitrary points of $\cS$, not at the key positions, so the value read at an atom must be defined everywhere; the observed $v_i$ are therefore interpolated into a field $V:\cS\to\R^d$ by partition-of-unity RBF interpolation with the same bandwidths:
\begin{equation}
\label{eq:field}
V(s) \;=\; \frac{\sum_{i=1}^n v_i\,\psi_\varepsilon(s-s_i)}{\sum_{i=1}^n \psi_\varepsilon(s-s_i)},
\qquad
\psi_\varepsilon(u) \;=\; e^{-\lVert u\rVert_\varepsilon^2/2}.
\end{equation}
The weights sum to one, so $V(s)$ is everywhere a convex combination of the observed values (a Nadaraya--Watson average).

\textbf{(iv) Output.} The L\'evy Attention output is the sample mean over the atoms $S_1,\dots,S_{Z_q}$ of $N_q$, with a mean fallback on the empty draw:
\begin{equation}
\label{eq:output}
\mathbf{a}(q) \;=\;
\begin{dcases}
\frac{1}{Z_q}\sum_{j=1}^{Z_q} V(S_j), & Z_q \ge 1,\\[3pt]
\abar(q), & Z_q = 0,
\end{dcases}
\qquad
\abar(q) \coloneqq \frac{\int_\cS V\,\lambda_q\,d\cL}{\Lambda_q}.
\end{equation}
Two remarks. Returning $\abar(q)$ on the empty draw is what keeps the operator exactly unbiased (\cref{thm:mean}) and leaves no query unanswered; training and default inference never draw at all, substituting the deterministic mean $\abar(q)$, which has exact gradients, throughout (\S\ref{sec:modes}). The fallback value $\abar(q)$ is the \emph{mollified cosine attention}: as $\varepsilon_t,\varepsilon_v\to 0$ it is exactly the discrete cosine-kernel attention $\sum_i \kappa(q,k_i)v_i/\sum_i\kappa(q,k_i)$ for pairwise-distinct key positions.

\subsection{Discretisation and implementation}
\label{sec:disc}

The operator of \S\ref{sec:stages} is defined through integrals over $\cS$. The implementation replaces them by sums over a fixed grid. This subsection gives the full recipe and states exactly what the grid changes and what it preserves.

\textbf{Grid.} $\cS=[0,1]^2$ is partitioned into $L = L_t\times L_v$ axis-aligned cells of equal Lebesgue measure $\Delta = 1/L$, with centres $s^*_1,\dots,s^*_L$. The resolution is tied to the bandwidths, $L_t=\lceil 1/\varepsilon_t\rceil$ and $L_v=\lceil 1/\varepsilon_v\rceil$, so one cell spans about one bandwidth per axis. The defaults are $\varepsilon=(1/16,1/8)$, giving the $16\times 8$ grid. \S\ref{sec:exp_abl} sweeps this choice.

\textbf{Discrete mollifier.} The bump of \cref{eq:mollifier} is evaluated at the cell centres and renormalised over the grid, per key:
\begin{equation}
\label{eq:dmoll}
\delta^\Delta_\varepsilon(s^*_l - s_i) \;=\; \frac{e^{-\lVert s^*_l - s_i\rVert_\varepsilon^2/2}}{\Delta \sum_{l'=1}^{L} e^{-\lVert s^*_{l'} - s_i\rVert_\varepsilon^2/2}},
\qquad
\Delta\sum_{l=1}^{L} \delta^\Delta_\varepsilon(s^*_l - s_i) \;=\; 1 .
\end{equation}
Each key therefore carries exactly unit mass on the grid, wherever it sits relative to cell centres and boundaries, so no compatibility mass is lost to discretisation.

\textbf{Discrete intensity.} The per-cell intensities and their total are
\begin{equation}
\label{eq:dintensity}
\lambda_l \;=\; \tau\,\Delta \sum_{i=1}^n \kappa(q,k_i)\,\delta^\Delta_\varepsilon(s^*_l - s_i),
\qquad
\Lambda_q \;=\; \sum_{l=1}^{L} \lambda_l \;=\; \tau\sum_{i=1}^n \kappa(q,k_i).
\end{equation}
The second equality is exact by \cref{eq:dmoll}. The evidence meter on the grid equals its continuous counterpart; $\Lambda_q$ involves no approximation. On the deterministic mean path (\S\ref{sec:modes}) $\tau$ cancels, since only the normalised intensity $\lambda_q/\Lambda_q$ enters, so it is a non-parameter there; in the sampled modes it sets the expected atom count. For the signals below it rescales $\Lambda_q$ uniformly across queries, which leaves the $1/\Lambda_q$ ranking exactly unchanged and the $\sighat$ ranking unchanged wherever $\varphi(\Lambda)\approx1/\Lambda$, the regime every experiment here operates in. Its effect on absolute scale is absorbed by the one-time validation calibration of \S\ref{sec:setup}.

\textbf{Discrete value field.} \Cref{eq:field} is evaluated at the cell centres, giving one vector $V_l = V(s^*_l)$ per cell through the same partition-of-unity weights. Cells far from every key inherit a blend of the nearest values; they also receive vanishing attention intensity, so they do not influence the output.

\textbf{Sampling on the grid.} The grid sampler draws a vector of independent per-cell counts, and \cref{eq:count} becomes their sum:
\begin{equation}
\label{eq:counts}
N_l \;\sim\; \mathrm{Poisson}(\lambda_l)\ \ \text{independently for}\ l=1,\dots,L,
\qquad
Z_q \;=\; \sum_{l=1}^{L} N_l \;\sim\; \mathrm{Poisson}(\Lambda_q).
\end{equation}
The sampled output and the fallback of \cref{eq:output} become
\begin{equation}
\label{eq:dout}
\mathbf{a}(q) \;=\;
\begin{dcases}
\frac{1}{Z_q}\sum_{l=1}^{L} N_l\,V_l, & Z_q \ge 1,\\[3pt]
\abar(q), & Z_q = 0,
\end{dcases}
\qquad
\abar(q) \;=\; \sum_{l=1}^{L} p_l\,V_l, \quad p_l = \frac{\lambda_l}{\Lambda_q}.
\end{equation}
The discretisation does not disturb the statistics of the operator. Conditional on the total count $Z_q$, the vector of cell counts is multinomial with probabilities $(p_1,\dots,p_L)$: each atom lands in cell $l$ independently with probability $p_l$. This is the discrete form of the Poisson process's conditional property.

\textbf{What discretisation costs.} Two approximations enter. The evidence carries none, by \cref{eq:dintensity}. The output $\abar(q)$ is a coarse-grained rather than discrete-kernel attention, and that mollification gap is the one we bound (\S\ref{sec:theory}, \cref{app:gap}); the grid's own effect on accuracy we measure instead, by sweeping its resolution (\S\ref{sec:exp_abl}).

\textbf{Engineering details.} Heads are folded into the batch; each head learns its own channel map $w_v$, so key positions differ per head, and the three signals of \cref{eq:signals} are computed per head and reported as head means. Queries are decoded in chunks to bound the kernel-matrix memory. Empty-history samples, present in USHCN month-chunks, receive one neutral token so that the encoder's masked attention stays defined.

\subsection{Two modes, one operator}
\label{sec:modes}

\textbf{Deterministic mean mode (default; training and bulk inference).} Training a network through a sampled operator is normally the painful part. One must either differentiate through the draw with a relaxation such as Gumbel--softmax, or accept high-variance gradients; either way the loss becomes noisy. Here none of this is needed. Wherever the network would use the sampled output $\mathbf{a}(q)$, we substitute its expectation $\abar(q)$ from \cref{eq:dout}. At the layer this substitution is exact, not an approximation: the mean identity of \S\ref{sec:theory} (\cref{thm:mean}) guarantees $\E[\mathbf{a}(q)]=\abar(q)$. And $\abar(q)$ is an ordinary deterministic expression, an $O(Ld)$ weighted sum with exact gradients, at the same cost as one sampled output. Training therefore involves no sampling at all: no relaxation, no train/inference distribution gap, no sampling variance in the loss. At inference the same substitution applies. If only a point prediction is wanted, output $\abar(q)$ and never draw an atom. We call this deterministic route the \emph{mean path}, as opposed to the sampled path of \cref{eq:dout}.

\textbf{What is then still random?} The randomness is not discarded; it is \emph{metered}. The same forward pass emits, per query at $O(L)$ extra cost on top of a shared $O(Ld)$ precompute,
\begin{equation}
\label{eq:signals}
\underbrace{\Lambda_q}_{\text{evidence}}, \qquad
\underbrace{\tr\SigV(q) = \sum_l p_l\,\lVert V_l\rVert^2 - \lVert\abar(q)\rVert^2}_{\text{disagreement}\; (p_l = \lambda_l/\Lambda_q)}, \qquad
\sighat(q) = \sqrt{\tr\SigV(q)\,\varphi(\Lambda_q)},
\end{equation}
where $\varphi(\Lambda)=\E[\mathbf{1}\{Z\ge 1\}/Z]\simeq 1/\Lambda$ (\cref{thm:variance}). $\sighat(q)$ is precisely the root-mean-square deviation the sampled operator \emph{would} exhibit, obtained without drawing a single sample. Three clarifications. $\varphi(\Lambda)$ is the expected reciprocal of the number of atoms being averaged, zero on the empty draw where the output sticks to the mean, and behaves as $1/\Lambda$ once atoms are plentiful. $\sighat(q)$ is one scalar per query, the root expected squared Euclidean deviation of the layer's $d$-dimensional output; calibrating it against held-out targets turns it into a scale for the head's scalar prediction (\S\ref{sec:exp_dist}); per-coordinate spread, if it is wanted, needs $K$ draws of the layer output. And no loss term ever touches these signals; they are read-outs of quantities the trained kernel already computes. We call them the \emph{free signals}: no extra pass, no extra parameters, no training change. Together they are the model's \emph{uncertainty interface}.

\textbf{$K$-draw mode (on-demand error bars).} When a predictive distribution is required, the kernel $\kappa(q,k_i)$, the cell intensities $\lambda_l$ and the value field $V_l$ are computed once, and only the $O(L)$ count vector $(N_1,\dots,N_L)$ is redrawn $K$ times; the $K$ outputs propagate through the (small) remainder of the network. The empirical spread converges to $\sighat(q)$ at the layer; in operation count this yields end-to-end error bars at $1$ encoder pass $+\,K$ re-runs of the decode step (\S\ref{sec:cost}). \S\ref{sec:exp_cohort} measures this.

\subsection{Computational cost}
\label{sec:cost}

For $m$ queries, $n$ keys, $L$ grid cells and head dimension $d$, the operator shares softmax attention's $O(mnd)$ kernel cost and adds three grid terms: $O(nLd)$ to build the value field (once per head), $O(mnL)$ to assemble the cell intensities, and $O(mLd)$ for the output reduction.

Since $L$ is constant in $n$, the total remains $O(mnd)$ asymptotically, and the premium is confined to the one layer the operator replaces (\S\ref{sec:exp_dist}). Drawing $K$ samples redraws only the length-$L$ count vector, so it costs $O(KmL)$ to redraw and $O(KmLd)$ to re-average, plus $K$ passes of the small decode head, not $K$ passes of the network. Our implementation re-runs the whole network per draw; \cref{tab:cohort} prices it.

\section{Theory}
\label{sec:theory}

This section establishes the claims the construction rests on: the mean path is unbiased (\cref{thm:mean}), and the pre-sample signals are exactly the deviation of the sampled operator (\cref{thm:variance}). Throughout, expectations are over $N_q$ conditioned on inputs. The results are stated in grid form, with continuum counterparts in parentheses, and their proofs use two distributional facts: conditional on the total count $Z_q$, the cell counts are multinomial with probabilities $p_l=\lambda_l/\Lambda_q$ (\S\ref{sec:disc}), and $Z_q$ itself is Poisson$(\Lambda_q)$, which is what turns the per-count statements into statements about the evidence. They therefore hold for the per-cell sampler of \cref{eq:dout} exactly as stated, and no separate discrete analysis is needed. Full proofs are in \cref{app:proofs}.

\begin{theorem}[Mean identity]
\label{thm:mean}
For every query with $\Lambda_q>0$, \;$\E[\mathbf{a}(q)] = \abar(q)$.
\end{theorem}

The proof is the conditional property of Poisson processes: given $Z_q=z\ge 1$ the atoms are i.i.d.\ with density $\lambda_q/\Lambda_q$, so every conditional mean equals $\abar(q)$, and the $\{Z_q{=}0\}$ branch returns $\abar(q)$ by construction. Two consequences: the mean path used for training is unbiased for the sampled operator, and the deviation identity below is centred at the trained prediction itself.

\begin{theorem}[Exact variance identity]
\label{thm:variance}
For every query with $\Lambda_q>0$, let $\SigV(q) = \sum_l p_l V_l V_l^\top - \abar(q)\abar(q)^\top$ with $p_l=\lambda_l/\Lambda_q$ be the covariance of the value field under the normalised intensity (in the continuum, the covariance of $V(S)$ under $S\sim\lambda_q/\Lambda_q$). Then
\begin{equation}
\label{eq:variance}
\E\bigl\lVert \mathbf{a}(q)-\abar(q)\bigr\rVert^2 \;=\; \tr\SigV(q)\cdot \varphi(\Lambda_q),
\qquad\text{so}\qquad
\sighat(q) \;=\; \sqrt{\tr\SigV(q)\,\varphi(\Lambda_q)}.
\end{equation}
Both factors are reductions over quantities the forward pass has already produced, and the third is a scalar function of the first:
\begin{align}
\label{eq:howto}
\underbrace{\Lambda_q \;=\; \sum_{l=1}^{L}\lambda_l \;=\; \tau\sum_{i=1}^{n}\kappa(q,k_i)}_{\text{evidence}},
\qquad
&\underbrace{\tr\SigV(q) \;=\; \sum_{l=1}^{L} p_l\lVert V_l\rVert^2 \;-\; \lVert\abar(q)\rVert^2}_{\text{disagreement}},
\nonumber\\[2pt]
\varphi(\Lambda) \;=\; \E\!\left[\tfrac{\mathbf{1}\{Z\ge1\}}{Z}\right]
\;=\; e^{-\Lambda}\!\int_{0}^{\Lambda}\!\frac{e^{t}-1}{t}\,dt
&\;=\; \frac{1}{\Lambda} + O(\Lambda^{-2}),
\qquad Z\sim\mathrm{Poisson}(\Lambda).
\end{align}
So the deviation scale is one square root of two grid reductions, and it factorises as
\[
\sighat(q) \;\approx\; \frac{\sqrt{\text{disagreement}}}{\sqrt{\text{evidence}}}.
\]
\end{theorem}

\Cref{thm:variance} is an identity rather than a bound. It also fixes the direction of the evidence, and repairs an intuitive but wrong signal. Because $\sighat$ falls as the evidence $\Lambda_q$ rises, the uncertainty signal the evidence induces on its own is the \emph{reciprocal} $1/\Lambda_q$; this is the quantity scored as ``evidence only'' in the tables below, while $\Lambda_q$ itself remains the evidence meter. On its own it ignores the values entirely, so a query attending to a hundred \emph{identical} values reports the same uncertainty as one attending to a hundred conflicting ones, though its true variance is zero. The disagreement factor is therefore not optional, and \S\ref{sec:exp_unc} confirms it.

One caution about the shape of $\varphi$. It is not monotone: it vanishes at $\Lambda=0$, rises to a maximum of about $0.52$ near $\Lambda\approx1.5$, and decays as $1/\Lambda$ beyond. The reading ``less evidence, larger $\sighat$'' therefore applies for $\Lambda_q\gtrsim 2$; below that, the $Z_q{=}0$ fallback dominates and the sampled operator sticks to the mean, so $\sighat$ shrinks again. Every experiment in this paper operates far above this region: mean evidence ranges from $62$ to $632$ even under the coverage stress test of \cref{fig:stress}, and on the PhysioNet interpolation test set the smallest evidence over all $65{,}976$ queries is $28$.

\paragraph{Mollification bias.} How far is $\abar(q)$ from the discrete cosine-kernel attention it replaces? A standard mollification argument (\cref{app:gap}) bounds the gap by a Gaussian tail in $\tilde r_{\min}$, the minimum key separation measured in the mollifier metric $\lVert\cdot\rVert_\varepsilon$: the two operators agree up to a gap that decays like $n\,e^{-\tilde r_{\min}^2/32}$, so separation must grow like $\sqrt{\log n}$ bandwidths for it to bite. Our operating points do not enter it: bandwidths stay wide, so $\abar$ acts as a smoother over neighbouring observations. Many keys share a cell, and $\abar$ is best read as a \emph{coarse-grained} attention in its own right. Whether coarse-graining costs accuracy is an empirical question, answered by the matched softmax control in \S\ref{sec:exp_acc}.

\textbf{What the identities certify.} The sampled operator, exactly, around the trained prediction: sampling is replaced by a closed form, and the structure of the signals is derived rather than chosen. What they do not settle is predictive error; that transfer is a measurement, and \S\ref{sec:exp_unc} makes it --- with a shape the decomposition predicts: disagreement should out-rank the combined $\sighat$ where errors are dominated by value conflict, and evidence should start contributing where they come from missing mass. Both happen. The identity also separates their uses: $\tr\SigV$ is a per-query ordering, while $\sighat$ is the layer's own deviation, in the units of an output, so one global affine map turns it into an interval scale.

\section{Related Work}
\label{sec:related}

\textbf{Attention for irregular time series.} mTAN \citep{shukla2021mtan} attends from reference time points with learned time embeddings; ContiFormer \citep{chen2023contiformer} pairs attention with ODE dynamics; GraFITi \citep{yalavarthi2024grafiti} and t-PatchGNN \citep{zhang2024tpatchgnn} treat IMTS forecasting as graph interaction, the latter defining the benchmark protocol we adopt. Under the t-PatchGNN protocol all of them are scored as point predictors and report no per-query reliability; where a probabilistic variant exists (mTAN's VAE, Latent-ODE, CRU's filtered state), its spread costs repeated latent draws and the protocol never reports it. Our operator is closest in spirit to mTAN (mTAND in \cref{tab:acc}; continuous-time kernel smoothing of observations) but replaces the deterministic average with a metered random measure whose moments are the uncertainty interface. \Cref{tab:acc} puts the two under the benchmark's own protocol, where our L\'evy model is the more accurate on all three datasets.

\textbf{Uncertainty in deep models.} Ensembles \citep{lakshminarayanan2017} and MC dropout \citep{gal2016} pay $K$ models or $K$ passes; CSDI \citep{tashiro2021csdi} obtains calibrated distributions for TS imputation at ${\sim}100$ diffusion samples per query; heteroscedastic heads \citep{kendall2017} train an uncertainty output; SNGP \citep{liu2020sngp} obtains single-pass uncertainty from a distance-aware GP output layer, and DDU \citep{mukhoti2023ddu} from a post-hoc feature-space density on a spectrally-normalised backbone. Our signal differs in provenance: it is not \emph{trained} to predict error and requires no architectural constraint. It is the closed-form second moment of the attention operator itself, with \cref{thm:variance} tying it to the output distribution by identity rather than by fit; both factors are derived quantities, not learned outputs. Evidential deep learning \citep{sensoy2018evidential,amini2020evidential} makes the closest terminological contact: it too reports single-pass uncertainty through an \emph{evidence} quantity, but there the evidence is the learned output of a head trained under a bespoke loss, whereas $\Lambda_q$ is the kernel's partition function, read off the pass rather than fitted.

\textbf{Stochastic and continuous attention.} Bayesian Attention Modules \citep{fan2020} sample attention weights from a learned distribution, estimating uncertainty by repeated passes; continuous-domain attention \citep{martins2020} moves the index space to the continuum but keeps a deterministic density; Performer-style random features \citep{choromanski2021} use sampling to \emph{approximate} softmax cheaply. Our Poisson construction uses randomness for neither regularisation nor approximation: the count channel makes the partition function observable ($\Lambda_q$) and the value spread quantifiable ($\tr\SigV$), \emph{before} any sample is drawn.

\textbf{Probabilistic models with native IMTS uncertainty.} GRU-ODE-Bayes \citep{debrouwer2019gruode} propagates a filtering posterior through continuous-time dynamics; Gaussian-process adapters and multi-task GPs \citep{li2016gpadapter,futoma2017mgprnn} carry a GP posterior over the imputed series into a downstream classifier; neural processes and their attentive variant \citep{garnelo2018np,kim2019anp} deliver uncertainty at arbitrary index points through a latent process. These models obtain uncertainty by committing to a probabilistic architecture; our interface instead retrofits an attention layer and leaves the backbone untouched: its signals certify the operator exactly rather than approximating a posterior. \S\ref{sec:exp_unc} runs the family's most directly comparable representative, the deterministic-path variant TNP-D \citep{nguyen2022tnp}, under our own protocol and reports where it wins (error ranking) and where it loses: accuracy, and the predictive distribution that follows from it (\S\ref{sec:exp_dist}). On the kernel-attention side, random-feature attentions \citep{peng2021rfa} use sampling, and linear attentions \citep{katharopoulos2020linear} deterministic feature maps, to approximate softmax. DUQ \citep{vanamersfoort2020duq} joins SNGP and DDU in the deterministic single-pass family. $\beta$-NLL \citep{seitzer2022bnll} targets the accuracy cost of heteroscedastic training; \S\ref{sec:exp_unc} runs it on both suites. Our split-conformal wrapper of \S\ref{sec:exp_dist} is textbook split conformal \citep{lei2018conformal}; the adaptive line that drops exchangeability to handle distribution shift \citep{gibbs2021conformal} is the branch it does \emph{not} belong to.

\section{Experiments}
\label{sec:experiments}

Every experiment serves one claim: \emph{the marginal cost of the uncertainty is zero, given the pass that already ran}. The qualifier is the operator swap itself, which is why accuracy comes first: the interface is only free if the predictions do not pay for it, and \cref{tab:acc} prices exactly that.

\subsection{Setup}
\label{sec:setup}

\textbf{Backbone.} A small transformer, trained from scratch for each dataset. Every observation becomes one token: a linear embedding of its value, Fourier features of its timestamp, and a learned per-variable vector. Three softmax self-attention blocks encode the $n$ tokens jointly. The encoded tokens become the keys and values of a single cross-attention decode layer, and that layer is the L\'evy layer. Each query is a $(t,\text{variable})$ pair, embedded like a token but without the value. Each token sits at its position $s_i$ of \cref{eq:positions}. A small MLP head turns the layer's output into the scalar prediction. The matched control swaps the one L\'evy layer for softmax cross-attention, under the same projections. Everything else is identical: same architecture, same recipe, ${\sim}0.7$M parameters. The two models differ by 516 parameters, the learned channel map of \cref{eq:positions}.

\textbf{Training.} Standard supervised regression, end to end, with masked MSE on the hidden targets. The loss is computed on the mean path: wherever the sampled output $\mathbf{a}(q)$ would appear, $\abar(q)$ appears instead (\S\ref{sec:modes}). Training therefore involves no sampling and no relaxation. Optimisation is Adam with early stopping on validation MSE at the benchmark's patience; the interpolation study instead uses AdamW for a fixed sixty epochs (\cref{app:details}). Poisson-sampled training exists only as an ablation arm (\S\ref{sec:exp_abl}). Full hyperparameters are in \cref{app:details}.

\textbf{Error-ranking metrics.} Uncertainty signals are scored by how well they rank per-query errors. Spearman $\rho$ between the signal and $|$error$|$ asks whether the signal orders queries the way their true errors do: $\rho=1$ is a perfect triage order, $0$ is no relation. AUSE \citep{ilg2018} asks how fast error leaves when the least-trusted queries are removed: discard queries in the signal's order and track the error of what remains (the sparsification curve), then measure the area between that curve and the one drawn by an oracle that discards by true error. Zero means the signal removes error as fast as knowing the errors would, and larger is worse. A query is one hidden (time, variable) target; both metrics pool all test queries of all records and variables into a single ranking, on the benchmark's normalised scale.

\textbf{Whose errors each signal ranks.} Both metrics are computed \emph{same-backbone}, since AUSE is defined against each model's own oracle. Signals read from one trained model --- ours and every post-hoc baseline --- therefore rank that same model's deterministic errors, which is what makes them directly comparable; baselines that predict differently by construction, such as the ensemble average or a trained head's own output, rank their own errors instead.

\textbf{Which metric decides.} \textbf{AUSE is the primary ranking metric of this paper}, because the decision the interface is built for is triage: discard the least-trusted queries and ask how fast error leaves. \textbf{Spearman $\rho$ is secondary}, reporting global monotonicity of signal against error. \textbf{CRPS} \citep[the continuous ranked probability score;][]{gneiting2007} \textbf{is the primary metric for predictive distributions} (\S\ref{sec:exp_dist}), and $\Delta$MAE and wall-clock are the cost axis against which every quality number is read.

\begin{figure*}[!tp]
\centering
\includegraphics[width=0.99\textwidth]{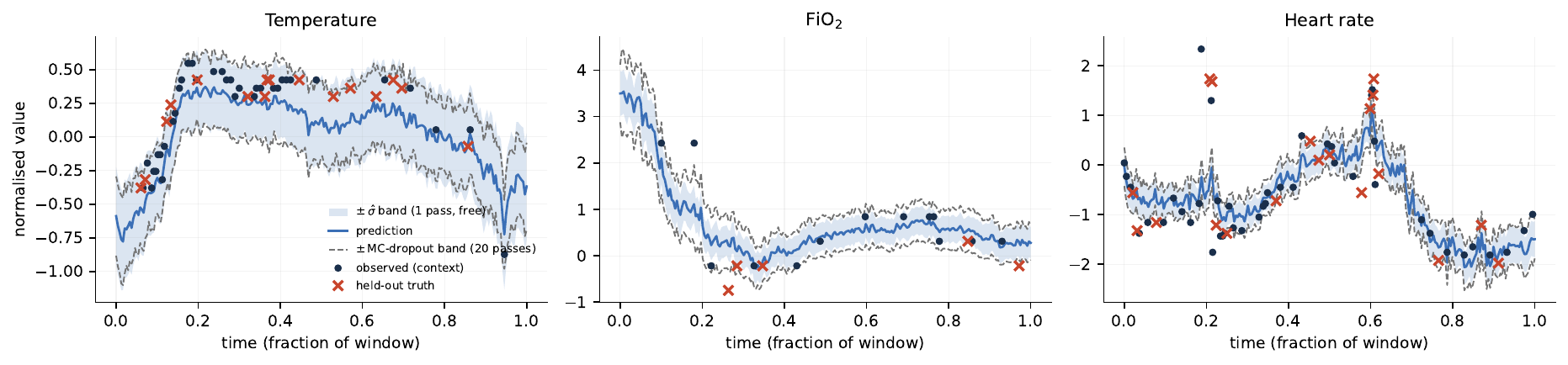}
\caption{The uncertainty interface in use: three variables from three held-out PhysioNet records. One deterministic forward pass of the seed-0 interpolation checkpoint produces the continuous prediction curve and the shaded band $a\,\sighat(q)+b$ at every point of the timeline (the validation calibration of \S\ref{sec:setup}). The band tightens where observations are dense and opens in the gaps. The dashed outline is the 20-pass MC-dropout band under its own validation-calibrated scale. The band is a one-sigma interval, not an envelope; \cref{tab:dist} measures coverage at the $80$, $90$ and $95\%$ levels.}
\label{fig:demo}
\end{figure*}

\textbf{Calibration.} Wherever ``val-calibrated'' appears, the interval scale is the affine map $a\,s(q)+b$ applied to whatever scale $s$ the estimator emits ($s=\sighat$ for ours), with $(a,b)$ fitted once on the validation targets, globally across variables, by a coarse grid search over the CRPS of the Gaussian predictive it defines. For a Gaussian, CRPS is closed-form, so neither the fit nor the reported score draws a sample. Nothing is ever fitted on test data. \Cref{fig:demo} previews what this buys: continuous confidence bands from the single deterministic pass, quantified in \S\ref{sec:exp_dist}.

\textbf{Protocol.} We evaluate on the t-PatchGNN benchmark \citep{zhang2024tpatchgnn} using its \emph{unmodified} public pipeline on its three openly-available datasets: PhysioNet 2012 (41 vars, 24h$\to$24h forecast), Human Activity (12 vars, 3000ms$\to$1000ms), and USHCN (5 vars, monthly chunks). MIMIC is excluded as it requires credentialed access under the PhysioNet Credentialed Health Data License. We report mean$\pm$std over 5 seeds unless stated otherwise, matching the benchmark's protocol; the transplant of \S\ref{sec:exp_transplant} and the rerun of \cref{tab:fore_bl} use 3, and every standard deviation in this paper is the unbiased sample estimator. A seed sets initialisation and training randomness; the benchmark's splits are fixed by its pipeline. In the interpolation study of \S\ref{sec:exp_unc} the seed additionally redraws the record split and the masking.

\textbf{Same-environment anchoring.} Published numbers can hide environment drift, and one benchmark metric (USHCN) lives on a dataset-specific scale. We therefore ran the benchmark's own t-PatchGNN, with its released code and canonical hyperparameters, \emph{in our environment} (last block of \cref{tab:acc}): it reproduces its published PhysioNet and Activity numbers to $0.7\%$ and $3.8\%$ respectively, validating the pipeline, and provides the reference value for USHCN. Every cross-environment comparison below is read against these anchors.

\subsection{Accuracy: what the operator swap costs}
\label{sec:exp_acc}

\begin{table}[t]
\centering\small
\setlength{\tabcolsep}{4.5pt}
\caption{IMTS forecasting under the t-PatchGNN protocol: MSE$\downarrow$, mean$\pm$std. $^{\dagger}$Quoted from \citet{zhang2024tpatchgnn} (Table~1; 5 seeds). $^{\ddagger}$Quoted from the same-protocol reruns of \citet{luo2025hipatch} (5 seeds). Anchor $=$ the benchmark's model run in \emph{our} environment; all rows measured by us use 5 seeds, matching the benchmark's seed protocol. Our rows share one untuned configuration across datasets. Percentages quoted against this table are computed from the unrounded means, not from the two-decimal cells. Bold: best per column; underline: second.}
\label{tab:acc}
\begin{tabular}{@{}lccc@{}}
\toprule
\textbf{Model} & \textbf{PhysioNet} ($\times10^{-3}$) & \textbf{Activity} ($\times10^{-3}$) & \textbf{USHCN} ($\times10^{-1}$) \\
\midrule
DLinear \citep{zeng2023dlinear}$^{\dagger}$ & 41.86 $\pm$ 0.05 & 4.03 $\pm$ 0.01 & 6.21 $\pm$ 0.00 \\
TimesNet \citep{wu2023timesnet}$^{\dagger}$ & 16.48 $\pm$ 0.11 & 3.12 $\pm$ 0.01 & 5.58 $\pm$ 0.05 \\
Crossformer \citep{zhang2023crossformer}$^{\dagger}$ & 6.66 $\pm$ 0.11  & 4.29 $\pm$ 0.20 & 5.25 $\pm$ 0.04 \\
Graph WaveNet \citep{wu2019graphwavenet}$^{\dagger}$ & 6.04 $\pm$ 0.28  & 2.89 $\pm$ 0.03 & 5.29 $\pm$ 0.04 \\
\midrule
GRU-D \citep{che2018grud}$^{\dagger}$ & 5.59 $\pm$ 0.09  & 2.94 $\pm$ 0.05 & 5.54 $\pm$ 0.38 \\
SeFT \citep{horn2020seft}$^{\dagger}$ & 9.22 $\pm$ 0.18  & 12.20 $\pm$ 0.17 & 5.80 $\pm$ 0.19 \\
Raindrop \citep{zhang2022raindrop}$^{\dagger}$ & 9.82 $\pm$ 0.08  & 14.92 $\pm$ 0.14 & 5.78 $\pm$ 0.22 \\
Warpformer \citep{zhang2023warpformer}$^{\dagger}$ & 5.94 $\pm$ 0.35  & 2.79 $\pm$ 0.04 & 5.25 $\pm$ 0.05 \\
mTAND \citep{shukla2021mtan}$^{\dagger}$ & 6.23 $\pm$ 0.24  & 3.22 $\pm$ 0.07 & 5.33 $\pm$ 0.05 \\
Latent-ODE \citep{rubanova2019latentode}$^{\dagger}$ & 6.05 $\pm$ 0.57  & 3.34 $\pm$ 0.11 & 5.62 $\pm$ 0.03 \\
CRU \citep{schirmer2022cru}$^{\dagger}$ & 8.56 $\pm$ 0.26  & 6.97 $\pm$ 0.78 & 6.09 $\pm$ 0.17 \\
Neural Flows \citep{bilos2021flows}$^{\dagger}$ & 7.20 $\pm$ 0.07  & 4.05 $\pm$ 0.13 & 5.35 $\pm$ 0.05 \\
GraFITi$^{\ddagger}$       & 5.11 $\pm$ 0.19  & 3.03 $\pm$ 0.14 & 5.07 $\pm$ 0.03 \\
t-PatchGNN$^{\dagger}$     & \underline{4.98 $\pm$ 0.08} & 2.66 $\pm$ 0.03 & 5.00 $\pm$ 0.04 \\
Hi-Patch$^{\ddagger}$      & \textbf{4.86 $\pm$ 0.03} & \textbf{2.57 $\pm$ 0.02} & 4.94 $\pm$ 0.05 \\
\midrule
t-PatchGNN (anchor, our env.) & 5.01 $\pm$ 0.06 & 2.76 $\pm$ 0.07 & 5.03 $\pm$ 0.07 \\
Backbone $+$ softmax decode (control) & 5.68 $\pm$ 0.27 & \underline{2.58 $\pm$ 0.02} & \underline{4.65 $\pm$ 0.06} \\
Backbone $+$ L\'evy decode (ours) & 5.80 $\pm$ 0.17 & 2.73 $\pm$ 0.04 & \textbf{4.64 $\pm$ 0.05} \\
\bottomrule
\end{tabular}
\end{table}

The load-bearing comparison in \cref{tab:acc} is between its last two rows: one backbone, one recipe, one parameter count, differing in a single layer. Against that matched control the L\'evy layer costs $2.0\%$ on PhysioNet, $5.6\%$ on Activity, and $-0.2\%$ on USHCN, a statistical tie ($4.64{\pm}0.05$ vs $4.65{\pm}0.06$). That is the accuracy cost of the interface: at most $5.6\%$, and on the sparsest dataset none at all.

Placed against the published field, the same two rows read as follows. On \textbf{PhysioNet} the model places fifth among the published IMTS methods, behind the same-environment anchor --- a gap the matched softmax control shares almost in full, so it is the backbone's and not the operator's. On \textbf{Activity} it places third among the published methods, between t-PatchGNN and Warpformer, and level with the same-environment anchor. On \textbf{USHCN} the L\'evy model attains a lower MSE than every published value in the column, Hi-Patch and t-PatchGNN included.\footnote{We flag this rather than celebrate it: USHCN is the benchmark's lowest-dimensional dataset (5 variables) and its metric lives on a dataset-specific scale; the same-environment anchor ($5.03$) validates the pipeline, and both decode layers share the value, so we read it as a backbone-family effect and draw no operator-level claim from it.} The control matches it, so we attribute the dataset to the backbone family rather than to the operator; what the operator adds there is the interface, at no accuracy cost at all.

What no row of \cref{tab:acc} reports is what the remaining sections measure: every prediction in our L\'evy row arrives with per-query evidence, disagreement, and a calibrated deviation scale, from the pass that produced it.

\subsection{A drop-in transplant into the benchmark model}
\label{sec:exp_transplant}

The drop-in claim is tested most directly by transplanting the operator into the state-of-the-art model itself. t-PatchGNN aggregates its $M$ patch tokens per variable with a time-blind linear layer and decodes each query time with an MLP; we replace only that output block, keeping the same MLP on top. Queries are the model's own time encodings, keys are the $M$ patch tokens positioned at their observation-time centroids, and attention runs per variable, so the index space is pure time. Parameter counts stay within $5\%$ of the original, a matched softmax transplant isolates the L\'evy construction, and every variant is trained \emph{from scratch} inside the benchmark's own unmodified loop. The comparison is three output blocks per dataset, each transplant trained at three seeds, all in the same batch.

\begin{table}[t]
\centering\small
\setlength{\tabcolsep}{5pt}
\caption{Transplanting the decode layer into t-PatchGNN: test MSE$\downarrow$ under its own protocol (mean$\pm$std over 3 seeds for the two transplants; the Original row is a single seed and carries no spread). The Original row is the unmodified model retrained in the same batch as the transplants. All three rows come from one batch, so the comparison is within-run throughout. Scales as in \cref{tab:acc}.}
\label{tab:transplant}
\begin{tabular}{@{}lccc@{}}
\toprule
\textbf{Output block} & \textbf{PhysioNet} ($\times10^{-3}$) & \textbf{Activity} ($\times10^{-3}$) & \textbf{USHCN} ($\times10^{-1}$) \\
\midrule
Original (time-blind linear $+$ MLP) & 5.11 & 2.70 & 5.01 \\
Softmax decode layer $+$ MLP & 5.22 $\pm$ 0.02 & 2.74 $\pm$ 0.13 & 5.02 $\pm$ 0.04 \\
L\'evy decode layer $+$ MLP & 5.17 $\pm$ 0.12 & 2.73 $\pm$ 0.06 & 5.02 $\pm$ 0.12 \\
\bottomrule
\end{tabular}
\end{table}

On every dataset the L\'evy transplant of \cref{tab:transplant} sits within its own three-seed spread of the single-seed original: the differences are at most $1.2\%$, smaller than the standard deviations printed beside them. It is also level with its own softmax transplant. The interface comes along at no extra cost. On sparse USHCN, the regime where the evidence factor carries information (\S\ref{sec:exp_unc}), the transplanted $\sighat$ retains predictive rank ($\rho=0.18{\pm}0.09$ over 3 seeds).

What a practitioner takes from this section is a recipe: an already-built IMTS model acquires the uncertainty interface for the cost of one output block and one retraining.

\subsection{Per-query uncertainty: how well the free signals rank errors}
\label{sec:exp_unc}

Throughout, the instruments under test are the three free signals of \cref{eq:signals}, read from the deterministic pass; everything they are compared against either ignores the trained model entirely or obtains its signal through extra passes, extra models, a post-hoc fit, or modified training. We are aware of no published error-ranking numbers for IMTS models, so every row here is measured by us on one backbone.

\begin{table}[t]
\centering\small
\setlength{\tabcolsep}{3pt}
\caption{Error ranking on the forecasting test sets (Spearman $\rho\uparrow$ / AUSE$\downarrow$; mean$\pm$std over 5 seeds), from the \emph{same single forward pass} as the predictions of \cref{tab:acc}; all three signals are free and available before the head has run.}
\label{tab:unc_fore}
\begin{tabular}{@{}lccc@{}}
\toprule
\textbf{Signal (1 free pass)} & \textbf{PhysioNet} & \textbf{Activity} & \textbf{USHCN} \\
\midrule
$1/\Lambda_q$ (evidence only) & $0.01{\pm}0.04$ / $0.59{\pm}0.03$ & $0.03{\pm}0.05$ / $0.61{\pm}0.05$ & $0.40{\pm}0.10$ / $0.43{\pm}0.06$ \\
$\tr\SigV(q)$ (disagreement only) & $0.22{\pm}0.10$ / $0.49{\pm}0.09$ & $0.09{\pm}0.06$ / $0.55{\pm}0.06$ & $0.47{\pm}0.11$ / $0.40{\pm}0.07$ \\
$\sighat(q)$ (ours, combined)        & $0.12{\pm}0.08$ / $0.51{\pm}0.05$ & $0.07{\pm}0.06$ / $0.56{\pm}0.06$ & $0.49{\pm}0.10$ / $0.37{\pm}0.06$ \\
\bottomrule
\end{tabular}
\end{table}

\textbf{The regime split.} \Cref{tab:unc_fore} shows the theory's decomposition living in the data. On \emph{dense} benchmarks (PhysioNet, Activity: many observations near every query) the evidence factor is uninformative ($\rho\approx0$) and \emph{disagreement} carries all the signal. On \emph{sparse} USHCN (monthly chunks with empty stretches) evidence becomes strongly informative ($\rho=0.40{\pm}0.10$ alone), and the combined $\sighat(q)$ ties the disagreement factor at the top on both metrics: $0.37{\pm}0.06$ against $0.40{\pm}0.07$ on AUSE, $0.49{\pm}0.10$ against $0.47{\pm}0.11$ on Spearman, with the per-seed difference changing sign. That swing, from $\rho\approx0$ on dense data to $0.40$ on sparse, is exactly what the decomposition predicts. Read \cref{eq:signals} accordingly \emph{as a ranker}: monitor $\tr\SigV$, read $\Lambda_q$ as the coverage meter, and take the combined $\sighat$ only where coverage is what varies. \S\ref{sec:exp_dist} reads it as a scale instead.

\textbf{The suite in the dense regime.} Hiding $30\%$ of the observed points of PhysioNet 2012 (set-a) yields about $66{,}000$ held-out targets per seed whose true values we know, and with them the room to run the established alternatives on one backbone (\cref{tab:unc_interp}). Read by what each row costs rather than by the table's blocks, the rows split in two. Among estimators that leave the trained predictor untouched, the free disagreement signal attains the best sparsification error measured (AUSE $0.438{\pm}.025$). One estimator ties it, a variance head fitted post hoc on the same frozen features ($0.441{\pm}.023$), and the head pays labels and a supervised fit for the tie. The other baselines in that tier rank behind both: twenty passes of MC dropout at $0.482$ (behind in every one of the five seeds), a DDU-style feature density at $0.476$, a last-layer Laplace posterior at $0.536$, every read-out of the matched softmax control, the closest being its own value dispersion at $0.471$, and every data-only heuristic, the closest at $0.606$. So $0.44$ is what reading a frozen model achieves on this dataset, across every estimator we measure. We call it the \emph{frozen-model ceiling} of this suite. The free signal reaches it with no training change, no extra pass and no head.

The other tier changes the model, and each route through it costs a training or a supervised fit. Five ensemble trainings buy a marginally better Spearman and a \emph{worse} AUSE ($0.461$); a Gaussian-NLL head buys a large ranking gain ($0.242$) at $+14.1{\pm}5.1\%$ MAE, which $\beta$-NLL halves without removing; and a matched Transformer Neural Process, probabilistic by construction, lands on our own NLL head to within noise ($0.247{\pm}.015$, losing to it in every seed) while costing $+25.4{\pm}7.0\%$ MAE.

\begin{table}[t]
\centering\small
\setlength{\tabcolsep}{4.5pt}
\caption{PhysioNet 2012 interpolation: per-query error ranking. Every row is mean$\pm$std over the same 5 seeds, one training per seed. The upper block holds estimators whose uncertainty is read out without any fit to labelled targets; the lower block holds estimators that fit theirs on labels, whether by changing the model, its training, or by fitting something new on a frozen one, each scored on its own errors. Bold: best AUSE in the upper block. ``Passes'': forward passes at inference. $\Delta$MAE: each row's test-MAE change against the point-head model (negative $=$ more accurate). $^{\S}$Data-only heuristics: observation pattern alone, no model and no forward pass. $^{\|}$Attention-weighted value dispersion, measured on the matched softmax control. $^{\P}$Last-layer Laplace \citep{daxberger2021laplace}. $^{**}$TNP-D \citep{nguyen2022tnp}, a Transformer Neural Process matched to our tokenisation, width, depth, optimiser and epoch budget. Recipes for all are in \cref{app:details}.}
\label{tab:unc_interp}
\begin{tabular}{@{}lcccc@{}}
\toprule
\textbf{Signal} & \textbf{Passes} & \textbf{Spearman} $\rho\uparrow$ & \textbf{AUSE}$\downarrow$ & \textbf{$\Delta$MAE} \\
\midrule
time gap, same variable$^{\S}$      & 0 & $0.019{\pm}.016$ & $0.698{\pm}.024$ & -- \\
time gap, any variable$^{\S}$       & 0 & $0.009{\pm}.002$ & $0.658{\pm}.014$ & -- \\
same-variable count$^{\S}$          & 0 & $0.043{\pm}.016$ & $0.606{\pm}.015$ & -- \\
windowed density$^{\S}$             & 0 & $0.038{\pm}.018$ & $0.621{\pm}.022$ & -- \\
softmax partition function (control) & 1 & $0.060{\pm}.019$ & $0.605{\pm}.013$ & -- \\
softmax attention entropy (control)  & 1 & $0.009{\pm}.022$ & $0.629{\pm}.016$ & -- \\
softmax value dispersion (control)$^{\|}$ & 1 & $0.138{\pm}.029$ & $0.471{\pm}.018$ & -- \\
$1/\Lambda_q$ (evidence only) & 1 & $0.067{\pm}.024$ & $0.610{\pm}.026$ & -- \\
$\sighat(q)$ (ours, combined)        & 1 & $0.124{\pm}.025$ & $0.536{\pm}.024$ & -- \\
$K$-draw spread ($K{=}8$)            & $1{+}8$ & $0.113{\pm}.025$ & $0.548{\pm}.023$ & -- \\
$\tr\SigV(q)$ (ours, disagreement)  & 1 & $0.213{\pm}.044$ & $\mathbf{0.438{\pm}.025}$ & -- \\
DDU-style feature density            & 1 & $0.180{\pm}.032$ & $0.476{\pm}.035$ & -- \\
last-layer Laplace$^{\P}$            & 1 & $0.123{\pm}.028$ & $0.536{\pm}.028$ & -- \\
MC dropout (20 passes)               & 20 & $0.182{\pm}.012$ & $0.482{\pm}.009$ & -- \\
deep ensemble ($5\times$ models)     & 5 & $0.225{\pm}.003$ & $0.461{\pm}.005$ & $-5.3{\pm}1.1\%$ \\
\midrule
\multicolumn{5}{@{}l}{\emph{Estimators that fit their uncertainty on labelled targets}} \\
post-hoc variance head (frozen backbone) & 1 & $0.216{\pm}.034$ & $0.441{\pm}.023$ & $0.0\%$ \\
heteroscedastic Gaussian-NLL head    & 1 & $0.487{\pm}.023$ & $0.242{\pm}.013$ & $+14.1{\pm}5.1\%$ \\
$\beta$-NLL head ($\beta{=}0.5$)         & 1 & $0.459{\pm}.006$ & $0.256{\pm}.002$ & $+7.2{\pm}2.3\%$ \\
TNP-D, natively probabilistic$^{**}$ & 1 & $0.485{\pm}.027$ & $0.247{\pm}.015$ & $+25.4{\pm}7.0\%$ \\
\bottomrule
\end{tabular}
\end{table}

\textbf{The suite in the sparse regime.} The free signals' strongest claims live where coverage varies, so the suite is rerun on USHCN interpolation; \cref{tab:unc_ushcn} reports the same-model signals with the sampling, density, and ensemble baselines, and the trained heads in its lower block. $\tr\SigV$ beats twenty-pass MC dropout on AUSE \emph{in every seed}, and its margin over dropout is five times the dense-data one ($0.341$ vs $0.560$ here, against $0.438$ vs $0.482$ on PhysioNet), while its Spearman leads in four seeds of five ($0.480{\pm}0.137$ vs $0.294{\pm}0.024$). It is a statistical tie with the DDU density (the seeds split $3$--$2$ on both metrics) and leads the last-layer Laplace in four seeds of five. The evidence factor, uninformative on dense data, here carries $\rho=0.34$ alone and carries $\sighat$ to $0.42$. The $K$-draw spread agrees with its closed form seed by seed, as \cref{thm:variance} demands; \cref{app:tests} checks the same identity directly. The data-only heuristics stay uninformative ($|\rho|\le0.05$ in every seed), so what the evidence factor reads on sparse data is not raw density.

Where error comes from missing coverage rather than value conflict, supervision of the representation pays: five trainings put the ensemble ahead on AUSE rather than behind it as on dense data ($0.302$ against $0.341$, inside the seed spread) at $\Delta$MAE $-1.1\%$, and the post-hoc variance head reaches $0.19$. A practitioner holding labels should spend them. The free signal is priced for the case where nobody is holding any: no labels, no fit, no extra pass, and available before the head has run. Its evidence factor is bimodal across seeds here: four of five give $\rho=0.35$ to $0.52$, one gives $-0.07$, and that single seed is the whole of the wide spread in the table.

TNP-D again out-ranks the free signal, at $+6.5{\pm}1.8\%$ MAE. But on this dataset it is itself out-ranked on AUSE in all five seeds by our own Gaussian-NLL head, and beaten, in four seeds of five on each metric, by the post-hoc variance head on the frozen model, which costs no accuracy at all. Committing to a probabilistic architecture yields less here than fitting a small head on a frozen deterministic one.

\begin{table}[t]
\centering\small
\setlength{\tabcolsep}{4.5pt}
\caption{USHCN interpolation in the sparse regime (30\% masking; 5 seeds, one training per seed; same protocol and metrics as \cref{tab:unc_interp}, over the same-model signals and the trained and sampling baselines). Spearman $\rho\uparrow$ / AUSE$\downarrow$, mean$\pm$std. No AUSE entry is bolded: the block minimum is a five-training ensemble, and among the one-pass rows the closest comparison (DDU) splits the seeds $3$--$2$. $\Delta$MAE is each row's test-MAE change against the point-head model, and (--) marks the upper-block rows that read that model's own predictions. $^{\S}$Data-only heuristics as in \cref{tab:unc_interp}: no model, no forward pass.}
\label{tab:unc_ushcn}
\begin{tabular}{@{}lcccc@{}}
\toprule
\textbf{Signal} & \textbf{Passes} & \textbf{Spearman} $\rho\uparrow$ & \textbf{AUSE}$\downarrow$ & $\Delta$\textbf{MAE} \\
\midrule
time gap, same variable$^{\S}$      & 0 & $0.009{\pm}.007$ & $0.728{\pm}.026$ & -- \\
time gap, any variable$^{\S}$       & 0 & $0.005{\pm}.006$ & $0.735{\pm}.034$ & -- \\
same-variable count$^{\S}$          & 0 & $-0.018{\pm}.025$ & $0.759{\pm}.027$ & -- \\
windowed density$^{\S}$             & 0 & $0.030{\pm}.011$ & $0.700{\pm}.015$ & -- \\
$1/\Lambda_q$ (evidence only)       & 1 & $0.341{\pm}.239$ & $0.506{\pm}.214$ & -- \\
$\sighat(q)$ (ours, combined)       & 1 & $0.415{\pm}.241$ & $0.412{\pm}.153$ & -- \\
$K$-draw spread ($K{=}8$)           & $1{+}8$ & $0.403{\pm}.234$ & $0.418{\pm}.145$ & -- \\
$\tr\SigV(q)$ (ours, disagreement)  & 1 & $0.480{\pm}.137$ & $0.341{\pm}.047$ & -- \\
DDU-style feature density           & 1 & $0.507{\pm}.041$ & $0.348{\pm}.033$ & -- \\
last-layer Laplace                  & 1 & $0.409{\pm}.066$ & $0.462{\pm}.090$ & -- \\
MC dropout (20 passes)              & 20 & $0.294{\pm}.024$ & $0.560{\pm}.050$ & -- \\
deep ensemble ($5\times$ models)    & 5 & $0.568{\pm}.030$ & $0.302{\pm}.026$ & $-1.1{\pm}1.2\%$ \\
\midrule
\multicolumn{5}{@{}l}{\emph{Estimators that fit their uncertainty on labelled targets}} \\
post-hoc variance head              & 1 & $0.660{\pm}.039$ & $0.191{\pm}.024$ & $0.0\%$ \\
Gaussian-NLL head                   & 1 & $0.653{\pm}.025$ & $0.187{\pm}.012$ & $+4.6{\pm}2.0\%$ \\
$\beta$-NLL head                    & 1 & $0.636{\pm}.081$ & $0.188{\pm}.045$ & $+9.3{\pm}6.6\%$ \\
TNP-D, natively probabilistic       & 1 & $0.619{\pm}.040$ & $0.210{\pm}.016$ & $+6.5{\pm}1.8\%$ \\
\bottomrule
\end{tabular}
\end{table}

\textbf{Beyond the two suites.} A three-seed rerun under the forecasting protocol extends the paid comparison to all three datasets (\cref{tab:fore_bl} in \cref{app:details}) and carries one fact: twenty dropout passes are an unstable instrument, tying, trailing or leading the free $\tr\SigV$ depending on dataset and metric, and failing outright on Activity ($\rho=-0.02{\pm}0.01$ at the default dropout rate, and negative at every rate of a single-seed sweep). The five-seed suites of \cref{tab:unc_interp,tab:unc_ushcn} supersede that rerun on the two datasets they share with it.

\subsection{Calibrated distributions, self-monitoring, and measured cost}
\label{sec:exp_dist}

So far the free signals have been used as ranks, and $\tr\SigV$ has carried that job. But $\sighat$ carries the units of a deviation, through \cref{thm:variance}, so it is the one we calibrate into a scale.

\textbf{What we compute.} \emph{(i) A predictive distribution.} For a query $q$ the network returns a scalar prediction $\mu(q)$, the head's output on the mean path $\abar(q)$; the same pass returns $\sighat(q)$ from \cref{eq:signals}. We read $\mu$ as a location and a rescaled $\sighat$ as a scale,
\begin{equation}
\label{eq:predictive}
p(y\mid q) \;=\; \mathcal N\bigl(\mu(q),\, s(q)^2\bigr), \qquad s(q) \;=\; a\,\sighat(q)+b .
\end{equation}
The rescaling is needed because $\sighat$ is the deviation of the layer's $d$-dimensional output, not of the scalar the head emits; the two scalars $(a,b)$ are fitted once on the validation split by the grid search of \S\ref{sec:setup}, and never on test.

A predictive law $F$ is scored against a realised target $y$ by its continuous ranked probability score,
\begin{equation}
\label{eq:crps}
\mathrm{CRPS}(F,y) \;=\; \int_{\mathbb R}\bigl(F(x)-\mathbf 1\{x\ge y\}\bigr)^2\,dx ,
\end{equation}
where $F$ is the predicted cumulative distribution function and $\mathbf 1\{\cdot\}$ the indicator. A point prediction is the degenerate law $F=\mathbf 1\{x\ge\mu\}$, for which the integral collapses to $|\mu-y|$: the CRPS of a point model is its MAE, which is what makes point and distributional rows directly comparable. For a Gaussian $F$ the integral is closed-form, so evaluating \cref{eq:crps} on \cref{eq:predictive} needs no sample.

\emph{(ii) Prediction intervals.} The \emph{coverage} of an interval is the fraction of test targets that fall inside it, and a level-$(1{-}\alpha)$ interval should have coverage $1-\alpha$. Reading the level off \cref{eq:predictive} gives the \emph{Gaussian interval} $G_\alpha(q)=\mu(q)\pm z_{1-\alpha/2}\,s(q)$, with $z_{1-\alpha/2}$ the standard normal quantile. Split conformal reads the multiplier off the data instead, assuming no shape at all. Over the $n_{\mathrm{cal}}$ validation queries $(q_i,y_i)$, score each by how many interval-widths it was off by, and take the corresponding upper quantile of those scores:
\begin{equation}
\label{eq:conformal}
R_i \;=\; \frac{|y_i-\mu(q_i)|}{s(q_i)}, \qquad
Q_\alpha \;=\; \text{the } \tfrac{\lceil (n_{\mathrm{cal}}+1)(1-\alpha)\rceil}{n_{\mathrm{cal}}}\text{ quantile of } \{R_i\}_{i=1}^{n_{\mathrm{cal}}}, \qquad
C_\alpha(q) \;=\; \mu(q)\,\pm\, Q_\alpha\, s(q).
\end{equation}
The conformal interval $C_\alpha(q)$ has coverage at least $1-\alpha$ in finite samples, whatever the shape of the errors, provided calibration and test queries are \emph{exchangeable}: their joint law is unchanged by reordering.

\emph{(iii) Self-monitoring.} The width $s(q)$ reacts to the input, not to past mistakes. Nothing is refitted at test time: the model is frozen and $(a,b)$ stay at their validation values. What does change is $\sighat(q)$, which is recomputed on every query. Hide observations and the evidence $\Lambda_q$ of \cref{eq:howto} falls; $\varphi(\Lambda_q)$ rises as $\Lambda_q$ falls; $\sighat(q)$ and hence $s(q)$ grow. The interval widens because the input got thinner, before any error has been observed.

\textbf{The measurements.} \Cref{tab:dist} scores \cref{eq:predictive} against the distributional baselines of \cref{tab:unc_interp}, each calibrated the same way. The last two rows are the same trained model read two other ways. Reported as a point prediction the model scores $0.317$, which is its MAE. Sampled, it saturates by $K{\approx}8$ and gets to $0.262$ at fifty draws. The Gaussian of \cref{eq:predictive} scores $0.244$ on those same predictions and draws nothing.

Three rows lose to it, and all three bought their scale by training: the $\beta$-NLL head at $0.255$, the Gaussian-NLL head at $0.269$, and TNP-D at $0.295$. Each loses in all five of its seeds. What they lose on is location, not shape. These same rows are among the sharpest error rankers in \cref{tab:unc_interp}, and its $\Delta$MAE column prices their ranking at $7$ to $25\%$ worse predictions. CRPS charges for that error directly, and a sharper ranking does not repay it. Two rows do better, and each names its price: the deep ensemble at $0.229$ for five trainings, the post-hoc head at $0.241$ for a set of labels and a fit on them. Five trainings and a label set are the two prices this paper measures everywhere else.

\begin{figure}[t]
\centering
\includegraphics[width=0.50\textwidth]{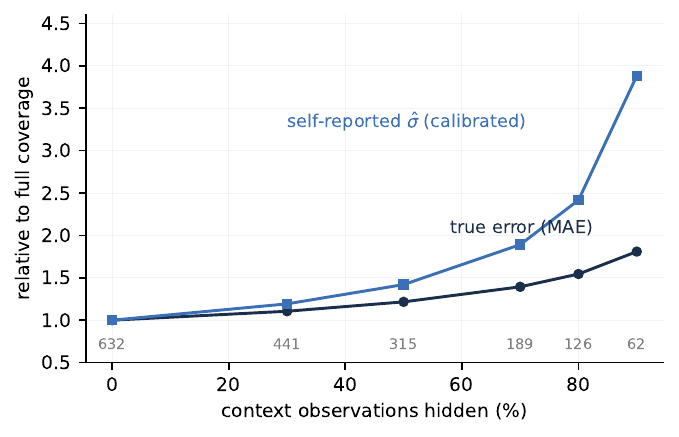}
\caption{Coverage-degradation stress test (PhysioNet interpolation, fixed targets). Both curves are normalised to their full-coverage value. Grey numbers give the mean evidence $\Lambda_q$ at each level. The model's self-reported deviation tracks its true degradation without ever seeing an error.}
\label{fig:stress}
\end{figure}

On coverage, $G_\alpha(q)$ is exact at the $80\%$ level and a few points short at $90\%$ and $95\%$, the signature of errors with heavier tails than a Gaussian; $C_\alpha(q)$ lands on the nominal level at all three, and its $95\%$ interval pays about a third more width for it. Under someone else's protocol the same read-out holds up: \cref{tab:csdi} runs it through the \emph{unmodified} CSDI imputation pipeline \citep{tashiro2021csdi}, where one deterministic pass places behind only the conditional diffusion the benchmark was built for, which needs one hundred samples per target. \Cref{fig:stress} exercises (iii): as we hide context, true error rises by $81\%$ and the self-reported $s(q)$ rises faster, driven by the evidence falling from $632$ to $62$.

The operator's own premium is one layer deep, $1.7\times$ a matched softmax cross-attention layer at our operating point ($n{=}512$, $m{=}128$, width $128$ over $4$ heads, $L{=}128$). \S\ref{sec:exp_cohort} prices all of it at deployment scale.

\begin{table}[t]
\centering\small
\setlength{\tabcolsep}{4pt}
\caption{Predictive distributions on PhysioNet 2012 interpolation. \textbf{Top:} CRPS$\downarrow$ at 30\% masking (mean$\pm$std over five seeds) for the distributional baselines of \cref{tab:unc_interp}; the last two rows are the same model read as a point prediction and by sampling (\S\ref{sec:modes}). Every row receives the same affine validation calibration; the raw column drops it and is empty for $\sighat$, which carries no units of the prediction. Bold: lowest calibrated CRPS. \textbf{Bottom:} empirical coverage and median interval width of the Gaussian and of its split-conformal wrapper, one checkpoint.}
\label{tab:dist}
\begin{tabular}{@{}lcc@{}}
\toprule
\textbf{Predictive distribution, 30\% masked} & \textbf{val-calibrated} & \textbf{raw} \\
\midrule
deep ensemble ($5\times$ models) & $\mathbf{0.229{\pm}.005}$ & $0.258{\pm}.005$ \\
post-hoc variance head (frozen backbone) & $0.241{\pm}.005$ & $0.248{\pm}.006$ \\
\emph{ours}: Gaussian on $\sighat$, 0 samples & $0.244{\pm}.006$ & -- \\
$\beta$-NLL head & $0.255{\pm}.012$ & $0.256{\pm}.011$ \\
Gaussian-NLL head & $0.269{\pm}.011$ & $0.271{\pm}.011$ \\
TNP-D, natively probabilistic & $0.295{\pm}.020$ & $0.298{\pm}.020$ \\
\midrule
\emph{ours}: deterministic mean (CRPS $=$ MAE) & $0.317{\pm}.007$ & -- \\
\emph{ours}: spread of $K$ draws, one seed, $K{=}1/8/50$ & $0.318/0.268/0.262$ & -- \\
\bottomrule
\end{tabular}\vspace{6pt}

\begin{tabular}{@{}lccc@{}}
\toprule
Nominal level & 80\% & 90\% & 95\% \\
\midrule
Gaussian coverage (\%) & 80.2 & 87.1 & 90.8 \\
Conformal coverage (\%) & 80.8 & 90.6 & 95.3 \\
\midrule
Gaussian median width & 0.87 & 1.12 & 1.33 \\
Conformal median width & 0.89 & 1.32 & 1.80 \\
\bottomrule
\end{tabular}
\end{table}

\begin{table}[t]
\centering\small
\caption{Our model under the unmodified CSDI imputation protocol (PhysioNet, $10\%$ missing, five folds). All rows use the quantile-CRPS estimator of \citet{tashiro2021csdi}, imported unmodified; for our Gaussian predictive its nineteen quantiles are available in closed form, so our row needs no sample. Our row is measured; the others are the values \citet{tashiro2021csdi} report under the same protocol.}
\label{tab:csdi}
\begin{tabular}{@{}lcc@{}}
\toprule
\textbf{Method} & \textbf{Samples per target} & \textbf{CRPS}$\downarrow$ \\
\midrule
CSDI (conditional diffusion) & 100 & $\mathbf{0.238{\pm}0.001}$ \\
\emph{ours}: Gaussian on $\sighat$ & \textbf{0} & $0.327{\pm}0.005$ \\
unconditional diffusion & 100 & $0.360$ \\
GP-based imputers & -- & $0.489$ and up \\
\bottomrule
\end{tabular}
\end{table}

\subsection{Screening an unseen cohort in seconds}
\label{sec:exp_cohort}

PhysioNet 2012 ships a second partition, set-b, that no training run, validation fit, or design decision in this paper ever touched; it is drawn from the same clinical distribution, and of its 4{,}000 records we keep the 3{,}383 with 20 to 512 time-varying observations, the loader's padding envelope. We stream it through the set-a-trained model as a deployment rehearsal. \textbf{One consumer GPU ranks the never-seen cohort by trust in 1.4 seconds; flagging the worst decile this way matches what 20-pass MC dropout delivers in 18.5 seconds.} \textbf{The cohort's full dense timeline, 35 million uncertainty-quantified queries, takes 6.8 seconds.} Every row of \cref{tab:cohort} is measured on the same RTX~3090.

\textbf{Protocol.} Each record conditions on 70\% of its observations; the remaining 30\% are hidden targets, over the challenge's 36 time-varying variables. One deterministic pass per batch returns all predictions, and its per-patient MAE is the error stream the single-model signals are scored on. A per-patient \emph{trust score} is the mean of a per-query uncertainty signal over the record's hidden-target queries, the same set whose MAE defines the error stream; sorting the cohort by it yields the intake list. Quality columns are measured over five same-recipe set-a trainings with fixed screening targets, reported as mean$\pm$std; wall-clocks come from one reported checkpoint.

\begin{table}[t]
\centering\small
\setlength{\tabcolsep}{4.5pt}
\caption{Cohort screening: the set-a-trained model on the untouched set-b partition (3{,}383 records; 30\% hidden targets per record, 373{,}713 queries), one RTX~3090. \textbf{Upper block:} patient-level triage quality, mean$\pm$std over five same-recipe trainings; wall-clock is the full screening pass on the reported checkpoint; bold marks the best wall-clock and the best mean of each quality column. Flagged decile: mean per-patient MAE of the 10\% least-trusted patients, as \% change against the cohort mean; larger means better ranking. $^{\P}$The ensemble is scored on its own averaged predictions (cohort MAE $0.305{\pm}0.002$ against the single model's $0.322{\pm}0.003$). $^{\P\P}$Post-hoc head: fitted per seed on frozen set-a features ($\Delta$MAE $=0$). $^{\|}$Softmax value dispersion: matched control, single run (cohort MAE $0.306$). Neither is separately timed. \textbf{Lower block:} wall-clock for the dense continuum (36 variables $\times$ 288 times per record).}
\label{tab:cohort}
\begin{tabular}{@{}lcccc@{}}
\toprule
\multicolumn{5}{@{}l}{\textit{Screening: patient-level triage quality}}\\
\midrule
\textbf{Trust signal} & \textbf{Passes} & \textbf{Wall-clock} & \textbf{Spearman} $\rho\uparrow$ & \textbf{Flagged decile} \\
\midrule
$\tr\SigV$ (ours)   & 1 & \textbf{1.35 s} & $0.43{\pm}.08$ & $+29.9{\pm}2.8\%$ \\
$\sighat$ (ours)    & 1 & \textbf{1.35 s} & $0.18{\pm}.05$ & $+18.6{\pm}1.4\%$ \\
$1/\Lambda_q$ (ours) & 1 & \textbf{1.35 s} & $0.05{\pm}.03$ & $+9.4{\pm}0.7\%$ \\
MC dropout spread   & 20 & 18.5 s & $0.49{\pm}.01$ & $+29.6{\pm}0.7\%$ \\
Deep ensemble spread$^{\P}$ & 5 & 5.3 s & $\mathbf{0.52{\pm}.01}$ & $\mathbf{+32.9{\pm}0.9\%}$ \\
Post-hoc variance head$^{\P\P}$ & 1 & -- & $0.46{\pm}.05$ & $+30.5{\pm}1.8\%$ \\
Softmax value disp.\ (control)$^{\|}$ & 1 & -- & 0.28 & $+24.3\%$ \\
\midrule
\multicolumn{5}{@{}l}{\textit{Continuum: wall-clock for 35{,}074{,}944 uncertainty-quantified queries}}\\
\midrule
\multicolumn{2}{@{}l}{Free signals, 1 pass (ours)} & \textbf{6.8 s} & & \\
\multicolumn{2}{@{}l}{MC dropout, 20 passes} & 133.5 s & & \\
\multicolumn{2}{@{}l}{$K$-draw sampling, $K{=}50$} & 428.7 s & & \\
\multicolumn{2}{@{}l}{Deep ensemble, 5 passes (after 5 trainings, 14.7 min)} & 32.1 s & & \\
\bottomrule
\end{tabular}
\end{table}

\textbf{Triage quality.} Across five same-recipe trainings, the flagged decile carries $+29.9{\pm}2.8\%$ excess error under $\tr\SigV$ against $+29.6{\pm}0.7\%$ under twenty-pass dropout, a statistical tie at one fourteenth of the cost. Whole-cohort Spearman orders the estimators by what they cost: the five-model ensemble leads at $0.52{\pm}.01$ after five trainings, dropout follows at $0.49{\pm}.01$ after twenty passes, a label-fitted post-hoc head at $0.46{\pm}.05$, and one pass gives $0.43{\pm}.08$, matching or beating dropout on two of five trainings, though its spread across retrainings is the widest in the column. The softmax control's dispersion analogue reached $0.28$ on its single control run, well behind the free $\tr\SigV$. The regime story of \S\ref{sec:exp_unc} repeats at patient granularity: disagreement dominates, $\sighat$ is diluted by its evidence factor ($0.18{\pm}.05$), and evidence alone is uninformative ($0.05{\pm}.03$).

\textbf{The same deliverable, at cohort scale.} The continuum pass answers all $35{,}074{,}944$ queries in $6.8$ seconds, about $5.2$ million uncertainty-quantified predictions per second; the lower block of \cref{tab:cohort} prices the same deliverable for every alternative.

\subsection{Ablations}
\label{sec:exp_abl}

\begin{table}[t]
\centering\small
\caption{Ablations, all on PhysioNet interpolation. \textbf{Top --- index-space geometry} (MAE$\downarrow$; each row is compared against the matched softmax control of its own run, in row order $0.289$, $0.288$, $0.286$, $0.298$); bold: best MAE. Percentages quoted in \S\ref{sec:exp_abl} are computed from the unrounded values, not from the three-decimal cells. \textbf{Middle --- backbone size} (seed 0, one recipe): $\rho$ is Spearman between signal and error, PICP the fraction of targets inside the nominal-80\% interval under the per-size validation-calibrated scale; bold marks the best per row for MAE, $\rho$ and AUSE, while PICP is scored against the nominal $0.80$ so no entry is bolded. \textbf{Bottom --- grid resolution} (seed 0), $L_t\times L_v$ time cells $\times$ channel cells; bold: best AUSE.}
\label{tab:abl}
\begin{tabular}{@{}lcc@{}}
\toprule
\textbf{Index space $\cS$} & \textbf{Grid} & \textbf{MAE} \\
\midrule
time only & 8 & 0.386 \\
time only & 32 & 0.371 \\
time only & 64 & 0.365 \\
time $\times$ learned channel (ours) & $16\times8$ & \textbf{0.302} \\
\bottomrule
\end{tabular}
\vspace{7pt}

\begin{tabular}{@{}lcccccc@{}}
\toprule
 & 0.2M & 0.7M & 2.7M & 5.1M & 10.8M & 20.3M \\
\midrule
MAE & 0.407 & 0.305 & 0.285 & 0.273 & 0.263 & \textbf{0.251} \\
$\rho$ ($\sighat$) & $-0.03$ & 0.16 & 0.15 & 0.16 & 0.20 & \textbf{0.23} \\
AUSE ($\tr\SigV$) & 0.62 & 0.40 & 0.41 & 0.42 & \textbf{0.39} & 0.42 \\
PICP@80 & 0.79 & 0.79 & 0.79 & 0.79 & 0.78 & 0.78 \\
\bottomrule
\end{tabular}
\vspace{7pt}

\begin{tabular}{@{}lcccc@{}}
\toprule
\textbf{Grid $L_t\times L_v$} & $8\times4$ & $16\times8$ (default) & $32\times8$ & $16\times16$ \\
\midrule
MAE & 0.310 & 0.305 & 0.308 & 0.306 \\
AUSE ($\tr\SigV$) & 0.478 & 0.394 & 0.413 & \textbf{0.378} \\
\bottomrule
\end{tabular}
\end{table}

\textbf{The index space must spread co-temporal variables.} The top block of \cref{tab:abl} settles the geometry of the index space. With a \emph{purely temporal} index space the value field averages all variables observed near the same instant, and temporal resolution barely repairs it: MAE improves only from $0.386$ to $0.365$ across an $8\times$ range of cells, and each run stays ${\sim}30\%$ behind its own softmax control. A single learned channel coordinate (\cref{eq:positions}) brings the gap to $1.5\%$. For IMTS the index space is therefore (given time) $\times$ (learned channel): the data supplies the axis it can, and learning supplies the one it cannot.

\textbf{The interface at scale.} The free signals are not a small-model artefact. The middle block of \cref{tab:abl} grows the backbone from 0.2M to 20.3M parameters under one recipe (PhysioNet interpolation, seed 0). Below about 0.7M both accuracy and the signals collapse: the smallest model has $\rho\approx0$, and its calibration fits $a{=}0$, so its band is constant and its PICP entry measures the marginal error scale, not a per-query interval. From 0.7M upward, accuracy improves monotonically across the entire range, the error ranking of $\sighat$ trends upward with size (from $\rho=0.16$ to $0.23$, with a flat middle), the disagreement ranking stays stable, and the calibrated coverage (PICP) stays within a point of nominal through 5.1M parameters and within two points at the two largest sizes. The interface needs enough capacity to exist and then holds across the remaining $29\times$ span of parameters, as the size-independent identity of \S\ref{sec:theory} predicts.

\textbf{Sensitivity to the grid.} The discretisation is not a tuned parameter. The bottom block of \cref{tab:abl} sweeps it around the default $16\times8$ grid (PhysioNet interpolation, seed 0) and accuracy stays flat: MAE within $0.305$--$0.310$ across an $8\times$ range of cell counts. The ranking quality of the free signals is not monotone in cell count, with the best AUSE at $16\times16$. The rate $\tau$ is a non-parameter in the same sense, and more strongly: swept over $[0.1,\,2]$ on the same checkpoint, a $20\times$ range, it leaves test MAE at $0.3047$ to four decimals and moves the disagreement AUSE by $5{\times}10^{-4}$, as \S\ref{sec:disc} predicts for the mean path.

\textbf{Sampled vs.\ mean training.} The mean path costs nothing to take. Training on the Poisson-sampled path (straight-through) instead gives test MAE 0.307 against the mean path's 0.309, a $0.7\%$ difference well inside the seed spread and exactly what \cref{thm:mean} predicts; we therefore default to the mean path and keep its exact gradients for free. Each block of \cref{tab:abl} is a separate training run, and every comparison in this section is made within its own run.

\section{Discussion}
\label{sec:discussion}

L\'evy Attention turns the attention layer itself into the uncertainty interface: the partition function that softmax discards becomes an evidence meter, the value spread it never surfaces becomes a disagreement meter, and an exact variance identity welds the two into a per-query deviation scale available before sampling, or entirely without it. The empirical picture in \S\ref{sec:experiments} matches that construction from the single query up to the cohort. Better per-query error ranking than the free signal exists in the interpolation suites, and none of it comes there without changing the model or fitting something new. The free signal is not the best available estimate of error; it is the best available for nothing.

The practical consequence is a change of default. While uncertainty costs a second model, twenty extra passes, or a supervised probe fitted on labelled data, it remains a feature to be budgeted for and, in practice, dropped. Once it is a by-product of the pass that already ran, there is no longer a reason for a prediction at an arbitrary timestamp to arrive without it. The interface also answers for queries nobody asked to predict. $\sighat(q)$ is defined for any $q$, so one pass scores every candidate measurement location at once, where dropout would pay $K$ passes and an ensemble $K$ models.

\paragraph{Limitations.} The identities certify the operator's own sampling deviation, not predictive error; the transfer from one to the other is empirical, and \S\ref{sec:exp_unc} measures it rather than proves it. That measurement has edges: the evidence factor is uninformative wherever coverage is dense, on Activity none of the free signals ranks errors well (\cref{tab:fore_bl}), and on sparse USHCN the evidence factor is bimodal across seeds (\S\ref{sec:exp_unc}). The predictive distribution of \S\ref{sec:exp_dist} assumes a Gaussian shape whose tails the coverage table shows to be optimistic at the higher levels, and the conformal guarantee assumes exchangeability between calibration and test queries, so neither survives distribution shift unmodified. $\sighat$ is one scalar per query, the deviation of the layer's $d$-dimensional output, and needs its one-time validation calibration before it acts as a scale for the head's scalar prediction. Finally, the evidence is confined to a single cross-attention decode layer atop backbones of at most 20.3M parameters on three public IMTS benchmarks; the coverage table reports one checkpoint, and the ablations one seed.

\section*{Broader Impact Statement}
The cohort screening of \S\ref{sec:exp_cohort} is a deployment rehearsal, not a clinical tool. The trust score ranks records by how fragile the model's own answers are; it does not detect deterioration, and a well-covered record can still be mispredicted. Any clinical use would require prospective validation, and the intervals of \S\ref{sec:exp_dist} inherit the caveats stated there and in the Limitations above: Gaussian tails are optimistic at high levels, and the conformal guarantee does not survive distribution shift. More broadly, uncertainty estimates cheap enough to be always on can create false reassurance if consumers read them as error bounds; they are rankings and calibrated scales, and \S\ref{sec:exp_unc} measures exactly how far each can be trusted.

\section*{Reproducibility Statement}
Every number in this paper is measured by us, except the rows explicitly marked as quoted in \cref{tab:acc}, whose sources are cited; and the published CSDI-protocol baselines quoted in \cref{tab:csdi}. The benchmark pipelines we evaluate under are public and are imported unmodified: splits, normalisation, collate functions, and metrics are the original authors' code. All datasets are openly available (PhysioNet 2012 set-a and set-b, Human Activity, USHCN); MIMIC is excluded because it requires credentialed access. Seeds, splits, and hyperparameters are stated in \S\ref{sec:setup} and \cref{app:details}. On PhysioNet set-a, records denser than $512$ observations are subsampled with a per-file checksum seed, so the subsample is identical in every process and across runs. The complete code (operator, benchmark adapters, every experiment and figure script), the raw result files behind every table and figure, and the interactive in-browser demo of the trained model will be released. Every experiment in the paper runs on a single consumer GPU; the training suite costs under thirty GPU-hours, plus evaluation.

\bibliographystyle{tmlr}

\appendix

\section{Proofs}
\label{app:proofs}

\subsection{Proof of Theorem~\ref{thm:mean} (mean identity)}
By the conditional property of Poisson processes \citep{kingman1993}, given $Z_q=z\ge1$ the atoms $S_1,\dots,S_z$ are i.i.d.\ with density $f_q = \lambda_q/\Lambda_q$ on $\cS$. Hence
$\E[\mathbf{a}(q)\mid Z_q=z] = \E_{S\sim f_q}[V(S)] = \abar(q)$ for every $z\ge1$, independent of $z$. On $\{Z_q=0\}$ the output equals $\abar(q)$ by definition. Averaging over $Z_q$ gives $\E[\mathbf{a}(q)]=\abar(q)$. \qed

\subsection{Proof of Theorem~\ref{thm:variance} (variance identity)}
Conditional on $Z_q=z\ge1$, $\mathbf{a}(q)=\frac1z\sum_{j=1}^z V(S_j)$ with $V(S_j)$ i.i.d., mean $\abar(q)$ and covariance $\SigV(q)$ (the covariance of $V(S)$ under $S\sim f_q$; on the grid, of the $V_l$ under $p_l$). Therefore
\[
\E\bigl[\lVert\mathbf{a}(q)-\abar(q)\rVert^2 \,\big|\, Z_q=z\bigr] \;=\; \frac{\tr\SigV(q)}{z}, \qquad z\ge1,
\]
while the $\{Z_q=0\}$ branch contributes $0$ (the fallback returns $\abar(q)$ exactly). Taking expectation over $Z_q\sim\mathrm{Poisson}(\Lambda_q)$,
\[
\E\lVert\mathbf{a}(q)-\abar(q)\rVert^2 \;=\; \tr\SigV(q)\sum_{z\ge1}\frac{e^{-\Lambda_q}\Lambda_q^z}{z\,z!} \;=\; \tr\SigV(q)\,\varphi(\Lambda_q).
\]
For the asymptotics of $\varphi$: $\varphi(\Lambda)=\E[Z^{-1}\mathbf 1\{Z\ge1\}]$ is the truncated inverse moment of a Poisson variable, whose standard expansion $\Lambda^{-1}+\Lambda^{-2}+O(\Lambda^{-3})$ gives $\varphi(\Lambda)=1/\Lambda+O(1/\Lambda^{2})$ as $\Lambda\to\infty$. Every series term $e^{-\Lambda}\Lambda^z/(z\,z!)$ vanishes at $\Lambda=0$, so $\varphi(0^+)=0$; this is what makes $\varphi$ non-monotone, with the maximum $0.517$ at $\Lambda=1.503$ quoted in \S\ref{sec:theory}. \qed

\subsection{Mollification bias}
\label{app:gap}
\begin{proposition}[Mollification bias]
\label{prop:gap}
Let $\mathbf{a}^\star(q)$ be the discrete cosine-kernel attention and let
$\tilde r_{\min} = \min_{i\ne j}\lVert s_i - s_j\rVert_\varepsilon$ be the minimum key separation measured in the mollifier metric $\lVert u\rVert_\varepsilon^2 = u_t^2/\varepsilon_t^2 + u_v^2/\varepsilon_v^2$ of \S\ref{sec:stages}. For $\tilde r_{\min}\ge 4$,
$\lVert\abar(q)-\mathbf{a}^\star(q)\rVert \le C\,V_{\max}\,n\,e^{-\tilde r_{\min}^2/32}$ uniformly in $q$, where $V_{\max}=\max_l\lVert V_l\rVert$ (in the continuum $\sup_s\lVert V(s)\rVert$, which the partition-of-unity structure bounds by $\max_i\lVert v_i\rVert$) and $C$ absorbs the boundary renormalisation of the mollifier (bounded above and below at any fixed $\varepsilon$).
\end{proposition}

The separation must be measured in $\lVert\cdot\rVert_\varepsilon$, not Euclidean distance: with anisotropic bandwidths a Euclidean-separated pair can be arbitrarily close along the narrow axis, and the competing-key ratio below then fails. Proof sketch: decompose $\abar(q)-\mathbf{a}^\star(q)$ over per-key mollifier averages $\widetilde v_i - v_i$, where $\widetilde v_i = \int_\cS V(s)\,\delta_\varepsilon(s-s_i)\,d\cL(s)$ is the value field averaged against the mollifier centred on key $i$; split each integral at the $\lVert\cdot\rVert_\varepsilon$-ball of radius $\tilde r_{\min}/4$ around $s_i$; inside that ball every competing key $j$ satisfies $\lVert x-s_j\rVert_\varepsilon \ge 3\tilde r_{\min}/4$, so its mollifier ratio against key $i$ is at most $e^{-\tilde r_{\min}^2/4}$; outside, the Gaussian tail mass in the $\varepsilon$-metric is $O(e^{-\tilde r_{\min}^2/32})$; combine with the convexity of the kernel weights. \qed

\section{Experimental details}
\label{app:details}
\textbf{Backbone.} $d_{\text{model}}=128$, 3 encoder blocks, 4 heads, dropout $0.1$; decode layer as in \S\ref{sec:setup}; head MLP $128\to128\to1$. Adam, lr $10^{-3}$, early stopping patience 10 on validation MSE (the benchmark's protocol), batch sizes and histories per dataset as in the benchmark's canonical scripts (PhysioNet $24$h/32; Activity $3000$ms/32; USHCN $24$mo/192).
\textbf{Interpolation study.} PhysioNet 2012 set-a over the challenge's 36 time-varying variables, the general descriptors excluded (the forecasting pipeline of \cref{tab:acc} keeps its own 41-variable convention), 70/15/15 record split, per-variable z-scoring from train statistics, 30\% of observed points hidden per record, capped at the loader's 128-target envelope (resampled each epoch), 60 epochs, AdamW $3{\times}10^{-4}$; MC dropout uses 20 train-mode passes of the same checkpoint; the decode layer stays in mean mode, so the measured spread is dropout's alone and carries no Poisson sampling.
\textbf{Compute.} The training suite (more than a hundred runs) takes under thirty GPU-hours, plus evaluation; each run uses a single consumer GPU.
\textbf{Forecasting baseline rerun (\cref{tab:fore_bl}).} Non-finite outputs on empty-context extremes are zeroed, identically for both variants, on top of the neutral-token handling of \S\ref{sec:method}.
\textbf{TNP-D baseline (\cref{tab:unc_interp,tab:unc_ushcn,tab:dist}).} Deterministic Transformer Neural Process \citep{nguyen2022tnp}. Context tokens carry (time, variable, value); target tokens carry (time, variable) only. One masked transformer runs over the concatenation, with a mask that lets targets attend to the context and to themselves but never to another target, so no target sees another's answer. A two-layer head emits $(\mu,\sigma)$ per target with a softplus-plus-floor parameterisation, trained by Gaussian maximum likelihood. Width, depth, head count, dropout, optimiser, learning rate, batch size, epochs, splits and masking are ours, unchanged, so the comparison isolates the estimator. 
\textbf{CRPS comparison of predictive distributions (\S\ref{sec:exp_dist}).} Five seeds at 30\% masking. Per seed we train the point model, the Gaussian-NLL head, the $\beta$-NLL head and a five-member ensemble, and fit the post-hoc variance head on the point model's frozen features. Each row's predictive Gaussian uses that row's own location and scale, scored by the same closed-form CRPS of a Gaussian, with the affine scale $a\,s+b$ applied to that row's own scale $s$, fitted on validation by the same grid search as \S\ref{sec:setup}. Uncalibrated scores are also recorded for the rows whose scale carries the units of the prediction. TNP-D is scored by the same closed-form expression, with its own validation calibration. \textbf{Baseline suite.} Every row is oriented so that larger means less trusted, as $1/\Lambda_q$ is for the evidence (\S\ref{sec:theory}); counts and partition functions are therefore negated or reciprocated before scoring. The softmax read-outs come from the matched control's own attention weights $\alpha_i$ and its own errors: value dispersion $\sum_i\alpha_i\lVert v_i\rVert^2-\lVert\sum_i\alpha_i v_i\rVert^2$, the reciprocal partition function $1/\sum_i\kappa(q,k_i)$, and the entropy of $\{\alpha_i\}$. The four data-only heuristics read the observation pattern alone and vary only through the per-seed data split: time to the nearest same-variable observation, time to the nearest observation of any variable, the count of same-variable observations, and that count restricted to observations falling within one sixteenth of the record's time span of the query. The post-hoc variance head is trained per seed on that seed's frozen features, so its $\Delta$MAE is zero by construction. The deep ensemble trains five point-head models from different initialisations. The trained head replaces the final linear layer and its loss (Gaussian NLL). The DDU-style density fits a 20-component GMM on PCA-projected penultimate features of the trained point model. The $\beta$-NLL head reweights the per-point Gaussian NLL by $\sigma^{2\beta}$ with stopped gradient, $\beta=0.5$ \citep{seitzer2022bnll}. The last-layer Laplace fits a Gaussian posterior on the final linear layer by the generalised Gauss--Newton over the training queries, with homoscedastic noise from train residuals and prior precision grid-tuned on validation NLL. The CSDI-protocol run imports their public dataset, masking, folds, and quantile-CRPS code unmodified; our model trains per fold on their train split with self-masked targets, and the affine scale is fitted on their validation split by their own CRPS.

\begin{table}[h]
\centering\small
\setlength{\tabcolsep}{5pt}
\caption{Free against paid under the forecasting protocol, all three datasets. Superseded on PhysioNet and USHCN by the five-seed suites of \cref{tab:unc_interp,tab:unc_ushcn}; on Activity it is the paper's only comparison against a paid estimator. Spearman $\rho\uparrow$ except in the block labelled AUSE$\downarrow$; mean$\pm$std over 3 seeds. MC dropout ranks the L\'evy run's own error stream; the softmax rows rank the matched control's. \S\ref{sec:exp_unc} reads the numbers.}
\label{tab:fore_bl}
\begin{tabular}{@{}lccc@{}}
\toprule
\textbf{Signal} & \textbf{PhysioNet} & \textbf{Activity} & \textbf{USHCN} \\
\midrule
\multicolumn{4}{@{}l}{\emph{L\'evy model (free, 1 pass)}} \\
$\sighat(q)$ & $0.04{\pm}0.03$ & $0.07{\pm}0.05$ & $0.49{\pm}0.15$ \\
$\tr\SigV(q)$ & $0.21{\pm}0.05$ & $0.08{\pm}0.06$ & $0.55{\pm}0.12$ \\
$1/\Lambda_q$ & $-0.09{\pm}0.06$ & $0.04{\pm}0.03$ & $0.26{\pm}0.23$ \\
\midrule
MC dropout (20 passes) & $0.24{\pm}0.05$ & $-0.02{\pm}0.01$ & $0.52{\pm}0.09$ \\
\midrule
\multicolumn{4}{@{}l}{\emph{AUSE$\downarrow$ on the same runs}} \\
$\tr\SigV(q)$ & $0.497{\pm}.049$ & $0.545{\pm}.045$ & $0.332{\pm}.073$ \\
MC dropout (20 passes) & $0.405{\pm}.052$ & $0.658{\pm}.005$ & $0.407{\pm}.121$ \\
\midrule
\multicolumn{4}{@{}l}{\emph{Softmax control (own errors, 1 pass)}} \\
value dispersion & $0.13{\pm}0.11$ & $0.23{\pm}0.02$ & $0.22{\pm}0.17$ \\
partition function & $0.02{\pm}0.10$ & $0.08{\pm}0.01$ & $0.34{\pm}0.11$ \\
attention entropy & $0.07{\pm}0.07$ & $0.01{\pm}0.01$ & $-0.20{\pm}0.06$ \\
\bottomrule
\end{tabular}
\end{table}

\section{Calibration tests}
\label{app:tests}
Unit tests accompanying the code verify: (T1) the grid identity $\sum_l\lambda_l = \tau\sum_i\kappa(q,k_i)$, at a maximum relative error of $1.2{\times}10^{-7}$, which is float32 machine precision; (T2) the Monte-Carlo mean of the sampled operator converges to $\abar(q)$, matching it to under $1\%$ of its norm at $4{,}000$ draws; (T3) the ratio of empirical to predicted deviation ($\sighat$) stays within $2\%$ of one across queries at $4{,}000$ draws; (T4) identical values drive $\sighat\to0$ while $1/\Lambda_q$ remains bounded away from zero: the counterexample to evidence-only uncertainty.

\end{document}